\pdfoutput=1

\documentclass[11pt]{article}

\usepackage{booktabs}
\usepackage{array}
\usepackage{pifont}
\usepackage{xcolor}
\usepackage{longtable}
\usepackage{colortbl}
\usepackage{graphicx}
\usepackage{amsmath}
\usepackage{hyperref}
\usepackage{tcolorbox}
\usepackage{soul}
\usepackage{amsfonts}
\usepackage{bbm}

\newcommand{\cmark}{\textcolor{green!60!black}{\ding{51}}} 
\newcommand{\xmark}{\textcolor{red!80!black}{\ding{55}}}   

\usepackage[final]{acl}

\usepackage{times}
\usepackage{latexsym}

\usepackage[T1]{fontenc}

\usepackage[utf8]{inputenc}

\usepackage{microtype}

\usepackage{inconsolata}

\usepackage{graphicx}

\definecolor{VideoGreen}{HTML}{006400}     
\definecolor{QuestionBlue}{HTML}{0000FF}     
\definecolor{AnswerMagenta}{HTML}{FF00FF}     
\definecolor{CurrentOrange}{HTML}{FFA500}    

\newcommand{\name}{\textsc{Social Genome}}

\title{\name:\protect\\ Grounded Social Reasoning Abilities of Multimodal Models}

\author{
  Leena Mathur\textsuperscript{*\dag}\textsuperscript{1}, Marian Qian\textsuperscript{*}\textsuperscript{1}, Paul Pu Liang\textsuperscript{2}, Louis-Philippe Morency\textsuperscript{1}\\
  Carnegie Mellon University\textsuperscript{1}, Massachusetts Institute of Technology\textsuperscript{2}\\
   {\texttt{\{lmathur,marianq,morency\}@cs.cmu.edu, ppliang@mit.edu}}}

\begin{document}
\maketitle

\renewcommand{\thefootnote}{\fnsymbol{footnote}}
\footnotetext[1]{equal contribution, \dag corresponding author}
\renewcommand{\thefootnote}{\arabic{footnote}}

\begin{abstract}
Social reasoning abilities are crucial for AI systems to effectively interpret and respond to multimodal human communication and interaction within social contexts. We introduce \name, the first benchmark for fine-grained, grounded social reasoning abilities of multimodal models. \name\ contains 272 videos of interactions and 1,486 human-annotated reasoning traces related to inferences about these interactions. These traces contain 5,777 reasoning steps that reference evidence from visual cues, verbal cues, vocal cues, and external knowledge (contextual knowledge external to videos). \name\ is also the first modeling challenge to   study external knowledge in social reasoning. \name\ computes metrics to holistically evaluate semantic and structural qualities of model-generated social reasoning traces. We demonstrate the utility of \name\ through experiments with state-of-the-art models, identifying performance gaps and opportunities for future research to improve the grounded social reasoning abilities of multimodal models. 

\end{abstract}

\section{Introduction}
\label{sec:intro}
Humans rely on \textit{social reasoning} to interpret and navigate everyday interactions \cite{gagnon2021reasoning}. This form of reasoning is a core competency of social intelligence \cite{kihlstrom2000social, conzelmann2013new}, occurs with specialized neural and cognitive systems \cite{cao2024domain, read2013constraint}, and involves integrating information over time from multimodal behaviors such as gestures, language, and prosody \cite{morency2010modeling,read2014dynamic, liang2024foundations}. Multimodal cues are often \textit{fine-grained} (e.g., a fleeting glance),  \textit{interleaved} (e.g., a shrug followed by a sigh), and \textit{context-dependent}, requiring \textit{external knowledge} of contextual information to be interpreted accurately \cite{hechter2001social}. 

\begin{figure}[t!]
    \centering
\includegraphics[width=0.91\linewidth]{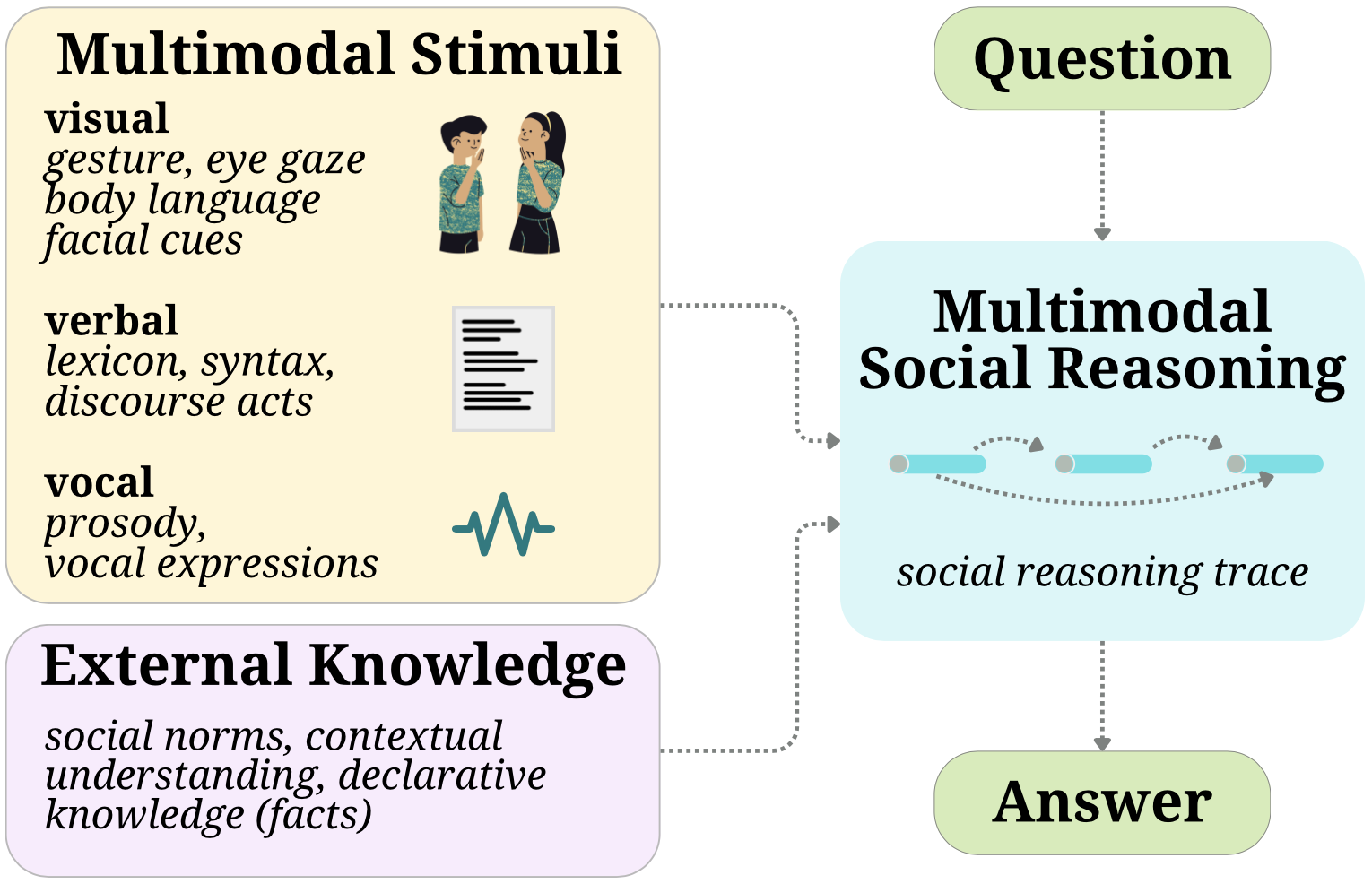}
    \caption{Reasoning over multimodal social interactions involves extracting, integrating, and referencing evidence from multiple behavioral modalities, as well as information from external knowledge.}
    \label{fig:sreason}
\end{figure}
\raggedbottom

Developing algorithms for multimodal social reasoning will be essential to advance artificial intelligence (AI) systems with social intelligence \cite{mathur-etal-2024-advancing}. 
When AI systems reason about human social interactions,  it is important for systems to have the ability to generate explanations with accurate, \textit{grounded} references to fine-grained multimodal behaviors and external  knowledge concepts informing inferences. Figure \ref{fig:sreason} visualizes these aspects of multimodal social reasoning. This capability is especially important for AI systems reasoning about interactions in high-stakes domains, such as healthcare and assistive robots.  

\begin{figure*}[t]
    \centering
    \includegraphics[width=0.93\linewidth]{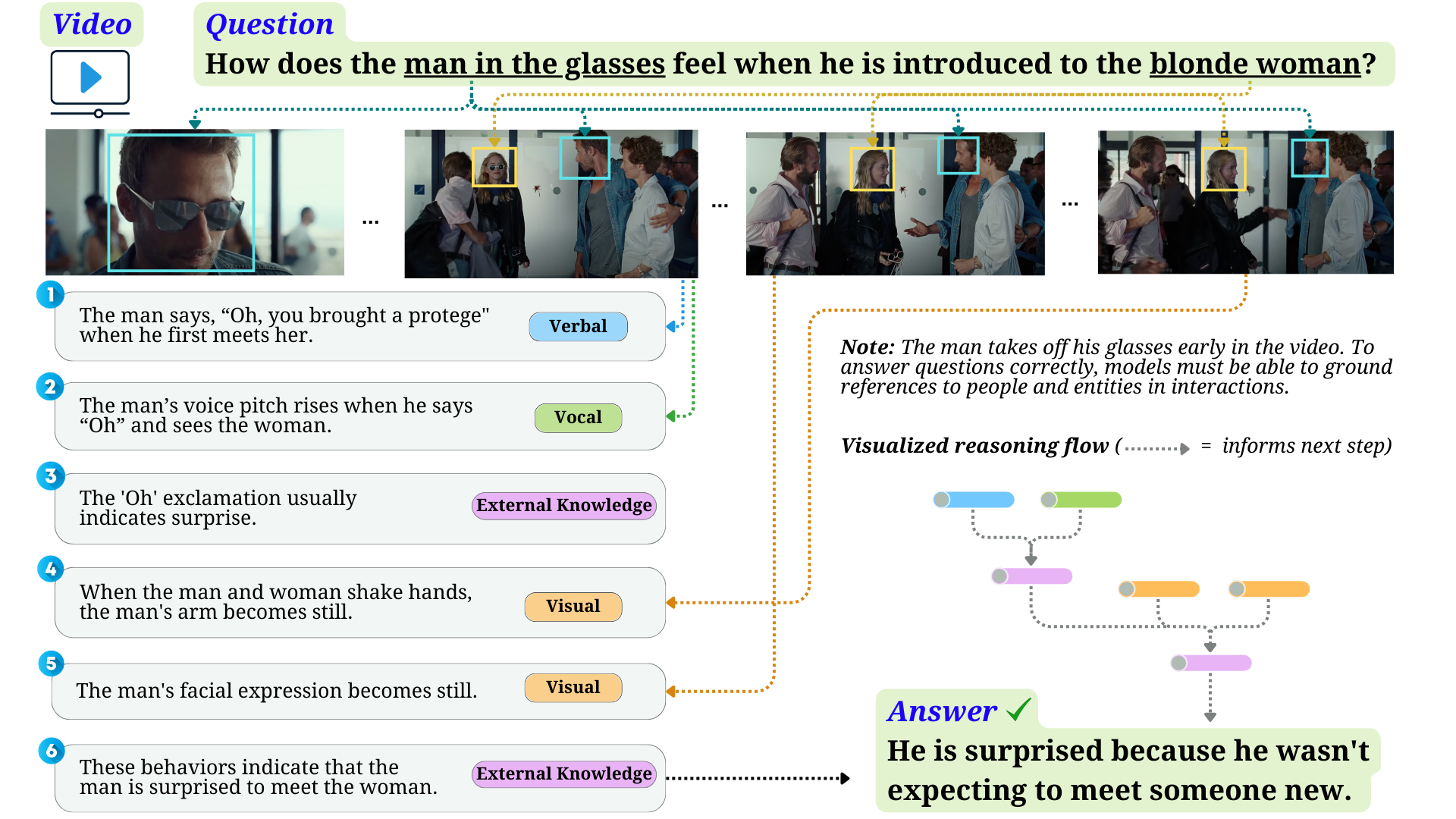}
    \caption{A sample reasoning trace from the \name\ benchmark. Reasoning traces in \name\ contain fine-grained, multimodal social cues and references to external knowledge informing the social inference. Social reasoning traces produced by humans can contain complex reasoning paths (sample visualized above) that reference and build upon multimodal evidence and external knowledge across  temporal segments of interactions.}
    \label{fig:overview}
\end{figure*}

 Progress towards improving multimodal social reasoning in models has been limited by a lack of evaluation tasks -- measuring a capability is an essential first step towards advancing it. To address this challenge, we introduce \textbf{\name}, the first benchmark for grounded multimodal social reasoning that includes 272 videos of face-to-face  interactions and 1,486 human-annotated reasoning traces explaining inferences about social information in these videos. Across these traces, \name\ contains 5,777 social reasoning steps. Each reasoning step is tagged with the modality of information being referenced: \textit{visual}, \textit{verbal}, and \textit{vocal} cues from social interactions in videos, and \textit{external knowledge} of contextual information that human annotators used to perform social inferences (information external to stimuli in videos). Reasoning traces in \name\ are \textit{dense} with references to over 11,000 entities (people, objects, concepts), over 5000 multimodal cues, and over 2,900 external knowledge observations. \name\ is the first social reasoning benchmark\footnote{\href{https://cmu-multicomp-lab.github.io/social-genome/}{\texttt{cmu-multicomp-lab.github.io/social-genome}}} that includes external knowledge and \textit{dense} reasoning traces. A sample human-annotated reasoning trace is visualized in Figure \ref{fig:overview}.

\begin{table*}[h]
  \centering
  
  \small
  \setlength\tabcolsep{4pt} 

   \begin{tabular*}{\textwidth}{@{\extracolsep{\fill}} l c c c c}

\textbf{Benchmark} &
\shortstack{\textbf{Social Reasoning}\\\textbf{Task Focus}} &
\shortstack{\textbf{Reasoning}\\\textbf{Traces}} &
\shortstack{\textbf{Fine-Grained}\\\textbf{Evaluation}} &
\shortstack{\textbf{External}\\\textbf{Knowledge}}\\
\midrule
\cellcolor{green!10}\textsc{Social Genome} (this paper)          & \cmark & \cmark & \cmark & \cmark \\
    \textsc{Social-IQ 1.0} \cite{zadeh2019social} & \cmark & \xmark & \xmark & \xmark \\
    \textsc{Social-IQ 2.0} \cite{wilf2023social}  & \cmark & \xmark & \xmark & \xmark  \\
    \textsc{TVQA} \cite{lei-etal-2018-tvqa}        & \xmark & \xmark & \xmark & \xmark \\
    \textsc{MovieQA} \cite{tapaswi2016movieqa}    & \xmark & \xmark & \xmark & \xmark  \\
    \textsc{MEmoR} \cite{shen2020memor}     & \xmark & \xmark & \xmark & \xmark  \\
    \textsc{Ego4D-Social} \cite{grauman2022ego4d}  & \xmark$^{\dagger}$ & \xmark & \xmark & \xmark \\
  \end{tabular*}

  \caption{\name\ enables the study of  fine-grained, grounded social reasoning in multimodal models.
  \cmark\ = provided; \xmark\ = not provided.  
  $^{\dagger}$Ego4D has a social signal perception task, which differs from a social reasoning task.}
  \label{tab:social_benchmarks}
\end{table*}

This paper defines metrics to holistically assess semantic and structural aspects of model-generated social reasoning traces. We demonstrate the utility of \name\ by using these metrics to distill insights regarding social reasoning capabilities and limitations in state-of-the-art (SOTA) models.  For example, we find that models struggle to perform well under both zero-shot and in-context learning (ICL) settings, demonstrating the significant challenge of building this understudied form of reasoning in models. Our findings contribute novel insights regarding gaps and opportunities for future research to improve grounded social reasoning abilities of multimodal models.

\section{Background}
\label{sec:related_work}

Prior research on social reasoning in models has primarily focused on the ability of models to interpret text-based social scenarios and perform question-answering (QA) tasks about characters' motivations, intents, and  actions; Social IQa remains a key unimodal benchmark in this area \cite{sap-etal-2019-social}. SOTA language models can accurately perform a majority of the inferences in Social IQa, but a gap remains between model and human performance \cite{sap-etal-2022-neural, shapira-etal-2024-clever}. SOTA models have also struggled with text-based QA tasks that probe  competencies relevant to social reasoning, specifically theory-of-mind to interpret the goals and beliefs of characters \cite{le2019revisiting, shapira-etal-2024-clever, ullman2023large, kim2023fantom}. Crowd-sourced knowledge bases of  norms \cite{forbes-etal-2020-social, ziems-etal-2023-normbank} have been useful to inform social reasoning research.

The ability of models to reason about multimodal social interactions, in particular \textit{face--to--face}, \textit{embodied}, \textit{real-world} social interactions, has been comparatively understudied. Key benchmarks include the video QA tasks of Social-IQ 1.0 \cite{zadeh2019social} and Social-IQ 2.0 \cite{wilf2023social};  both examine model QA accuracy when answering questions about social interactions in videos. SOTA models have struggled to perform well on Social-IQ 2.0 \cite{xie2023multi, pirhadi2023just, li2024llms, agrawal2024listen, chen2024through}. Prior focus on QA \textit{accuracy} to assess social reasoning ability has not enabled researchers to study the extent to which models can effectively reference \textit{fine-grained} multimodal cues and \textit{external knowledge} informing inferences. Models with high accuracy on QA tasks can perform poorly at generating valid or comprehensive reasoning traces \cite{jhamtani2020learning, gu2023digital}, motivating the creation of benchmarks with fine-grained evaluation  that goes beyond QA tasks. We introduce \name\ as the first benchmark to study grounded, fine-grained social reasoning in multimodal models. Table \ref{tab:social_benchmarks} summarizes the novelty of \name\, relative to prior human-centered video understanding tasks.  


\section{Building \name\ }
\label{sec:datset}

\subsection{Sourcing Seed Videos and Questions}
\label{subsec:seed_videos}
\name\ contains 272 seed videos and 1486 questions adapted from the \textsc{Social-IQ 2.0} dataset \cite{wilf2023social} (details in Appendix \ref{subsec:sourcing}). Videos include real-world face-to-face  interactions  (1 min per video, $\sim$4.5 hours); questions probe behaviors, emotions, and cognitive states of individuals and groups. \name\ introduces a new set of 1486 human reasoning traces with 5700+ steps that answer these questions.  

\subsection{Task Notation}
\label{subsec:task}
Given a video \( V \), a question \( Q \) about social interactions in the video, and answer options \( A = \{A_{\text{correct}}, A_{\text{incorrect}_1}, A_{\text{incorrect}_2}, A_{\text{incorrect}_3}\} \), a model performing the \name\ task must generate a reasoning trace \( R = \{e_1, e_2, \dots, e_n\} \), where each reasoning step \( e_i \) represents a single piece of evidence contributing toward the  social inference to select an answer \(A_{\text{a}}\) from \( A \). Each reasoning step \( e_i \) must be tagged with two attributes: (1) a modality tag \(m_i \in \{\textit{visual, verbal, vocal, } n/a\}\) indicating the communication modality of the evidence and (2) an external knowledge tag \(k_i \in \{\text{\textit{yes}, \textit{no}}\}\), indicating whether the evidence references external knowledge of contextual information. This task to generate \(\mathit{R}\) and answer question \textit{Q} 
evaluates a model’s ability to extract and reference multimodal aspects of human communication and knowledge informing social inferences. Given the input tuple \( (V, Q, A) \), each model performing the \name\ task will produce an output tuple \( (A_{\text{a}}, R) \). Metrics in \name\ study the social inference accuracy of \(A_{a}\) and the semantic and structural aspects of social reasoning in \(R\).

\subsection{Social Reasoning Trace Annotations}
\label{subsec:annotation}


\paragraph{Human Annotation} 
Given a video \( V \),  question \( Q \), and answer options \(A\), 
 annotators read \( Q \) and \( A\), watched  \( V\), and wrote reasoning trace \(R\). Annotations were collected with an IRB-approved Prolific study (details in Appendix \ref{subsec:annotation_appendix}).

\paragraph{Grounded and Fine-Grained Behaviors} Humans build upon low-level observations of fine-grained behaviors (e.g, shifts in body language) and high-level, top-down processing (e.g., implicit situational knowledge) when interpreting social scenes \cite{baird2001making, bodenhausen2013social}. Annotators were instructed to reference any \textit{low-level} and \textit{high-level} evidence they relied upon to answer questions: for example, low-level evidence might be "the woman takes a step back with her mouth wide open (visual cue)" and high-level evidence that interprets that low-level cue might be "the woman is surprised (external knowledge regarding how 'surprise' might manifest"). For each step, annotators tagged the modality referenced (visual, verbal, vocal, n/a). 

\paragraph{Grounded External Knowledge} Annotators tagged each reasoning step with \(yes\) or \(no\) to indicate whether external knowledge was referenced. External knowledge includes contextual norms, cultural expectations, and prior understanding of social commonsense \cite{forguson1988ontogeny} that goes beyond stimuli in the video. For example, if a man raises his arm and the annotator recognizes his movement as a "high five," the identification of the gesture is based on external knowledge. 

\paragraph{Ensuring Annotation Quality} 
Trained experts validated each Prolific annotation. They watched each video, read each QA tuple, and read the annotation to ensure that traces represented valid reasoning, had correct modality and external knowledge tags, and referred to relevant information. Cases of incomplete annotation or deviation from instructions were fixed (details in Appendix \ref{subsec:validation}).   

\subsection{Dataset Statistics}
\name\ contains 1486 human-annotated reasoning traces with 5,777 total steps, 3.89 $\pm$ 1.68 steps per trace (min of 1 step and max of 10 steps), 43 $\pm$ 26 words per trace, and 11$\pm$ 5 words per step.  Reasoning steps draw on multimodal evidence; 44\% of steps reference visual cues, 27\% reference verbal cues, and 17\% reference vocal cues. Overall, 77\% of traces reference at least one visual cue, 63\% reference at least one verbal cue, and 47\% reference at least one vocal cue.

External knowledge plays a critical role in \name: 51\% of reasoning steps referenced external knowledge, with each trace referencing an average of 2 pieces of external knowledge. With spaCy named entity recognition (NER) \cite{Honnibal_spaCy_Industrial-strength_Natural_2020}, we found 11,253 entities (people, objects, concepts) mentioned, with 7.6 unique entities and 2.23 emotions referenced
per reasoning trace, demonstrating the high \textit{density} of annotations. Additional details regarding dataset statistics are in Section \ref{sec:ethics} and Appendix \ref{subsec:data_info}.


\subsection{Social Reasoning Metrics and Statistics}
\label{subsec:metrics}
We develop metrics to evaluate \textit{semantic} and \textit{structural} aspects of social reasoning traces generated by models performing tasks in the \name\ benchmark. Collectively, these metrics reveal strengths and weaknesses in model social reasoning and multimodal grounding abilities and the extent to which model  traces differ from human reasoning. This multi-dimensional evaluation mitigates against models hacking individual metrics to achieve higher scores. For each sample, we compute the following metrics between model reasoning trace \(\mathit{R_M}\) with \(n\) steps $e_{1}...e_{n}$ and the corresponding human trace \(\mathit{R_H}\) with \(m\) steps $h_{1}...h_{m}$.

\paragraph{Accuracy} Measures accuracy of the model-generated answer by comparing it to ground truth. Human annotator accuracy on \name\ is 0.85 (Appendix \ref{subsec:human_acc}). Higher values indicate stronger social inference ability (max value is 1).  

\paragraph{Similarity-Trace (\(S_\text{trace}\))}  
Measures the high-level semantic similarity between \(\mathit{R_M}\) and \(\mathit{R_H}\):
\[
S_\text{trace} = 
\frac{\langle \mathbf{R_M}, \mathbf{R_H} \rangle}{\|\mathbf{R_M}\| \cdot \|\mathbf{R_H}\|}
\]
where \(\mathbf{R_M}\) is the aggregate embedding of evidence steps in \(\mathit{R_M}\) and \(\mathbf{R_H}\) is the aggregate embedding of evidence steps in \(\mathit{R_H}\). The embedding model \texttt{all-MiniLM-L6-v2} \cite{reimers-2019-sentence-bert}, selected for its efficiency and accuracy, was used to embed evidence steps for this and other semantic similarity metrics. Higher values indicate stronger alignment in semantic information between \(R_M\) and \(R_H\) (max value is 1).

\paragraph{Similarity-Step (\(S_\text{step}\))}  
Measures the fine-grained semantic similarity between \(\mathit{R_M}\) and \(\mathit{R_H}\). For each  step \(e_i\) in \(\mathit{R_M}\), the metric identifies its closest semantic step \(h_j\) in \(\mathit{R_H}\). The final metric is the mean of these maximum similarity values:  
\[
S_\text{step} = 
\frac{1}{n} \sum_{i=1}^n \max_{j} 
\frac{\langle \mathbf{e_i}, \mathbf{h_j} \rangle}{\|\mathbf{e_i}\| \cdot \|\mathbf{h_j}\|}
\]
where \(\mathbf{e_i}\) is the embedding of evidence step \(e_i\) and \(\mathbf{h_j}\) is the embedding of evidence step \(h_j\). Higher values reflect stronger alignment in semantic information between fine-grained steps of evidence in \(R_M\) and \(R_H\) (max value is 1).

\paragraph{Similarity-Num Steps (\(S_\text{num}\))}  
Measures the number of  steps in \(\mathit{R_M}\) with a  similarity above threshold \(\tau\), when compared to any step in \(\mathit{R_H}\):
\[
S_\text{num} = 
\sum_{i=1}^n \mathbbm{1} \left( 
\max_{j} \frac{\langle \mathbf{e_i}, \mathbf{h_j} \rangle}{\|\mathbf{e_i}\| \cdot \|\mathbf{h_j}\|} > \tau \right)
\]
where \(\mathbbm{1}(\cdot)\) is the indicator function and \(\tau=0.6\) (empirically selected). Higher values indicate more semantically-aligned evidence between \(R_M\) and \(R_H\) (max value is $n$).

\paragraph{DifferenceSequence (\(DS\))}  
Measures structural similarity between \(\mathit{R_M}\) and \(\mathit{R_H}\) using the respective modality sequences \(\mathit{S_M}\) and \(\mathit{S_H}\)  from the reasoning traces (e.g., ["visual", "external knowledge"]). A similarity score based on edit distance, adapted from the Levenshtein distance, is computed between \(\mathit{S_M}\) and \(\mathit{S_H}\) (Appendix \ref{subsec:diffseq}). Higher values of \(DS\) indicate greater structural similarity between \(R_M\) and \(R_H\) (max value is 1).

\paragraph{EmotionMetric}  
Measures the alignment of emotional content in \(\mathit{R_M}\) and \(\mathit{R_H}\). This metric extracts sets of emotions referenced by \(\mathit{R_M}\) and \(\mathit{R_H}\) (instructions in Appendix \ref{subsec:emotionmetric}) and computes the overlap in sets. Higher values can indicate stronger emotional alignment between \(\mathit{R_M}\) and \(\mathit{R_H}\) (max value is emotion set size in \(\mathit{R_H}\)). 

\paragraph{All Modality Steps}  
Measures the overlapping number of unique modalities in \(\mathit{R_M}\) and \(\mathit{R_H}\). Higher values indicate that \(\mathit{R_M}\) had more overlap with modalities referenced by \(\mathit{R_H}\) (max value is the number of unique modalities in \(\mathit{R_H}\)).

\paragraph{Visual Steps} 

Measures the number of steps in both \(\mathit{R_M}\) and \(\mathit{R_H}\) with \textit{visual} evidence, to
evaluate how closely \(\mathit{R_M}\) aligns with \(\mathit{R_H}\) (max value is number of visual steps in \(\mathit{R_H}\)).

\paragraph{Verbal Steps} 
Measures the number of steps in both \(\mathit{R_M}\) and \(\mathit{R_H}\) with \textit{verbal} evidence,
to evaluate how closely \(\mathit{R_M}\) aligns with \(\mathit{R_H}\) (max value is the number of verbal steps in \(\mathit{R_H}\)).

\paragraph{Vocal Steps}
Measures the number of steps in both \(\mathit{R_M}\) and \(\mathit{R_H}\) with \textit{vocal} evidence to evaluate
how closely \(\mathit{R_M}\) aligns with \(\mathit{R_H}\) (max value is the number of vocal steps in \(\mathit{R_H}\)).

\paragraph{External Knowledge Steps}
Measures the number of steps in both \(\mathit{R_M}\) and \(\mathit{R_H}\) with \textit{external knowledge} evidence, to 
evaluate how closely \(\mathit{R_M}\) aligns with \(\mathit{R_H}\) (max value is the number of external knowledge steps in \(\mathit{R_H}\)).

\paragraph{NumSteps (\(NS\))}  
Measures the absolute difference in the number of reasoning steps between \(\mathit{R_M}\) and \(\mathit{R_H}\). Lower values indicate stronger alignment in length between model and human chains (value of 0 indicates that \(\mathit{R_H}\) and \(\mathit{R_H}\) are the same length).

\begin{figure*}[t]
    \centering
\includegraphics[width=0.94\linewidth]{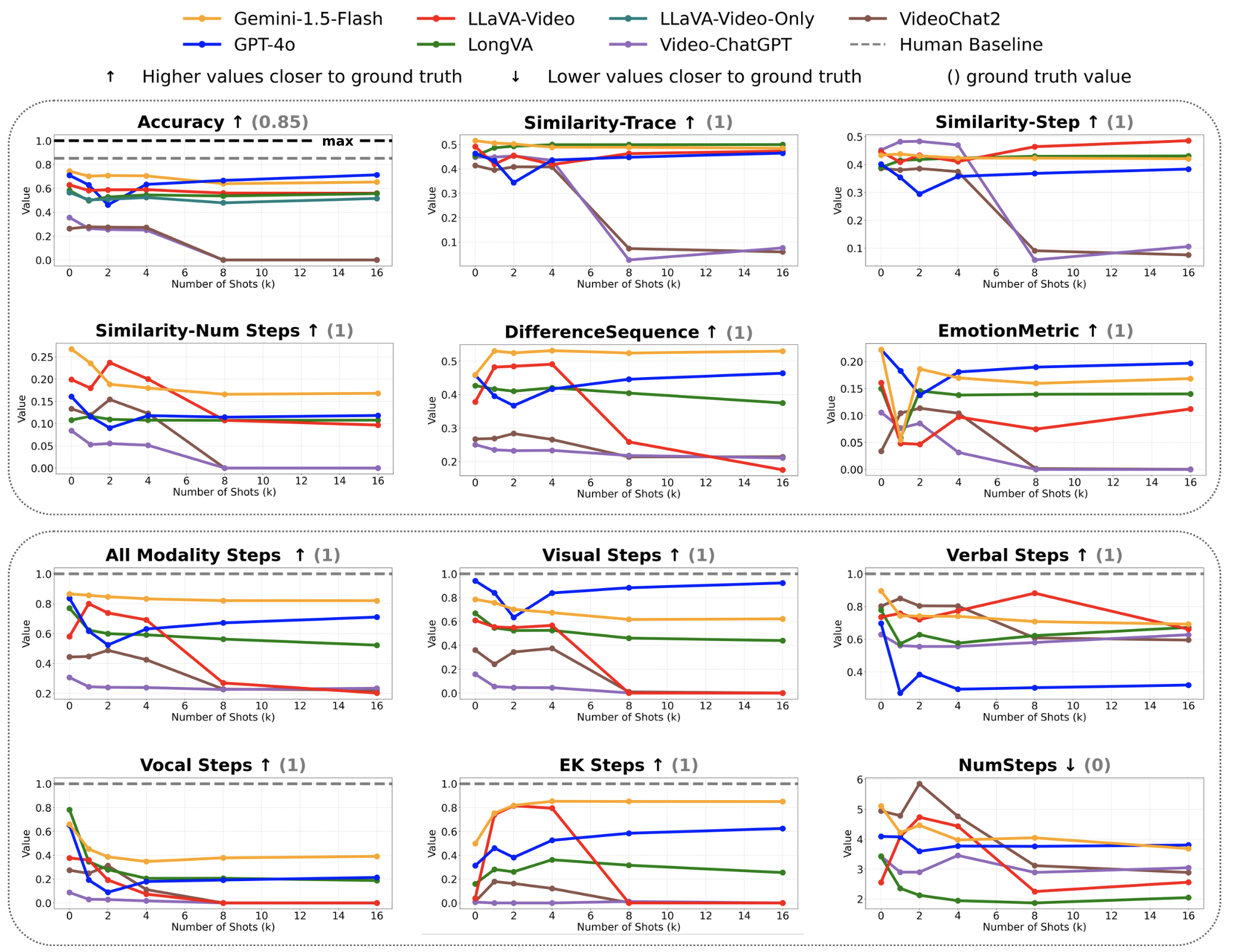}
    \caption{Performance of models across number of few-shot ICL samples ($k$) and ground truth noted in gray. The first six metrics focus on social inference accuracy, semantic similarity, and structural similarity between model and human reasoning traces. The final six metrics focus on fine-grained multimodal evidence and external knowledge referenced by models, in comparison to evidence referenced by humans. \textit{With these metrics, \name\ enables a holistic study of multimodal, grounded social reasoning.}}
    \label{fig:overall}
\end{figure*}

\subsection{\name\ ICL Training Set}
To create samples for ICL experiments, we randomly sampled 16 questions from unique videos in the \textit{training} set of \textsc{Social-IQ 2.0} and collected reasoning trace annotations in the same format as annotations in Section \ref{subsec:annotation}. ICL experiments in Section \ref{sec:experiments} are conducted by providing each model with different numbers of training samples $k \in \{0, 1, 2, 4, 8, 16\}$, before the model is given an input tuple ($V$, $Q$, $A$) and generates a sequence of tokens with an answer $A_a$ and reasoning trace $R$. 

\section{Social Reasoning Experiments}
\label{sec:experiments}

We use \name\ to study the performance of  multimodal video understanding models in fine-grained, grounded social reasoning. These models exhibited SOTA performance on video understanding tasks and take videos as input. We tested 2 closed-source models and 5 open-source models: Gemini-1.5-Flash \cite{team2024gemini}, GPT-4o \cite{gpt4o}, LLaVA-Video and LLaVA-Video-Only \cite{zhang2024video}, LongVA \cite{zhang2024longva}, Video-ChatGPT \cite{maaz2023video}, and VideoChat2 \cite{li2023videochat}. Models have different architectures, pretraining data, and fine-tuning tasks, and models generated reasoning traces and answers for all samples (details in Appendix \ref{sec:models-appendix}). LLaVA-Video-Only answered questions, but did not generate reasoning traces that could be studied; this model does not appear in trace-related metrics. 


\subsection{Quantitative Results and Insights}
\label{subsec:quant}

Figure \ref{fig:overall} visualizes each model's average performance for our 12 metrics\footnote{Visualized metrics are normalized to [0,1] by human baselines, except for those with an upper bound of 1 by definition (Accuracy, Similarity-Trace, Similarity-Step, DifferenceSequence) and absolute measures (NumSteps).}, with human baselines in gray. Results tables (Tables \ref{tab:accuracy-metrics-comparison}, \ref{tab:core-metrics}, \ref{tab:modality-metrics}) are in Appendix \ref{sec:models-appendix}. Key findings are discussed below. 

\paragraph{Social Inference Accuracy} \textit{Human social inference ability is substantially higher than all models}, as seen in results from the \textbf{Accuracy} metric. Gemini-1.5-Flash and GPT-4o achieve the highest accuracies of 74.4\% and 71.0\% respectively ($k$ = 0), approximately 10-15\% lower than human annotator accuracy (85.3\%), answering $\sim$30\% of questions incorrectly. \textit{Closed-source models outperform open-source models in social inference}. The highest-performing open-source model was LLaVA-Video at 62.9\% ($k$ = 0). Gemini-1.5-Flash and GPT-4o are much larger than open-source models, suggesting that increased model \textit{scale} is useful, but not sufficient, for social inference. Video-ChatGPT and VideoChat2 perform 30-40\% lower than other open-source models across all values of $k$. These models have the smallest context length, constraints which may influence performance. 

\textit{Social inference accuracy for models decreased as the number of few-shot ICL samples increased}, with the exception of GPT-4o, which demonstrated a slight improvement at $k$ = 16. Few-shot ICL conditions language models on tasks by providing examples of inputs and outputs \cite{liu2024incomplete} and can be viewed as a form of  \textit{inductive reasoning}, as the model is tasked with inferring generalizable rules from a set of examples. This technique has improved model reasoning abilities in domains such as mathematics and code generation \cite{dong2024survey,zhou2022teaching,patel2023evaluating}, which have \textit{explicit} rules and formal structure \cite{galotti1989approaches}. In contrast to these domains, social reasoning often operates with \textit{implicit} rules, less formal structure \cite{perkins1989reasoning} and ambiguity in premises \cite{mathur-etal-2024-advancing}. Our findings suggest that few-shot ICL may not be an effective approach to elicit multimodal social reasoning abilities. Additional experiments using \name\ samples as a form of supervision for models (discussed in Appendix \ref{sec:extra_experiments}) demonstrate that chain-of-thought prompting did not improve model accuracy, and models struggled to perform inferences relying on implicit and contextual knowledge. Our finding that few-shot ICL is insufficient to elicit multimodal social reasoning aligns with findings from unimodal experiments on \textsc{Social-IQa} and \textsc{ToMi} datasets \cite{le2019revisiting, sap-etal-2019-social,sap-etal-2022-neural,kim2023fantom}.

\paragraph{Semantic Alignment} \textit{It is challenging for models to generate social reasoning traces with high semantic alignment to human reasoning}, as seen in the results from the \textbf{Similarity-Trace}, \textbf{Similarity-Step}, and \textbf{Similarity-Num Steps} metrics. For Similarity-Trace, only Gemini-1.5-Flash achieved slightly above 50\% ($k$ = 0), and LongVA outperformed Gemini-1.5-Flash after $k$ = 4. For Similarity-Step, no models achieved above 50\%, but Video-ChatGPT outperformed all models for $k \in \{0, 1, 2, 4\}$ (5-12\% higher than GPT-4o and 2-5\% higher than Gemini-1.5-Flash). For Similarity-Num Steps, Gemini-1.5-Flash achieved the highest performance (27\%), far below  ground truth. Low performance can be explained by failure to reference cues that humans reference (``Fine-Grained Grounding" metrics below). Performance on these metrics did not improve as $k$ increased.

\paragraph{Structural Alignment} \textit{Model reasoning traces tend to reference multimodal evidence in different amounts and orders than humans}, as seen in the results from the \textbf{DifferenceSequence} metric. Model-generated sequences of multimodal evidence that were most structurally-aligned with human sequences were from Gemini-1.5-Flash (0.53 at $k$ = 4) and LLaVA-Video (0.49 at $k$ = 4), both far below the maximum alignment value (1). As $k$ increased, DifferenceSequence metric scores increased for Gemini-1.5-Flash, LLaVA-Video (up to $k$ = 4), and GPT-4o (after $k$ = 2). These findings suggest that structured samples of human social reasoning, as introduced by \name, can be useful when conditioning models to generate reasoning traces with more human-like structure.

\paragraph{Emotion Alignment} \textit{It is challenging for models to generate social reasoning chains with high  emotional alignment with human reasoning}, as seen in the results from the \textbf{EmotionMetric}. All models achieved less than 22\%, far below ground truth. While the highest scores were achieved by Gemini-1.5-Flash and GPT-4o at $k$ = 0, scores from several models steadily improved after $k$ = 2. 

\begin{figure*}[t]
    \centering
    \includegraphics[width=0.92\linewidth]{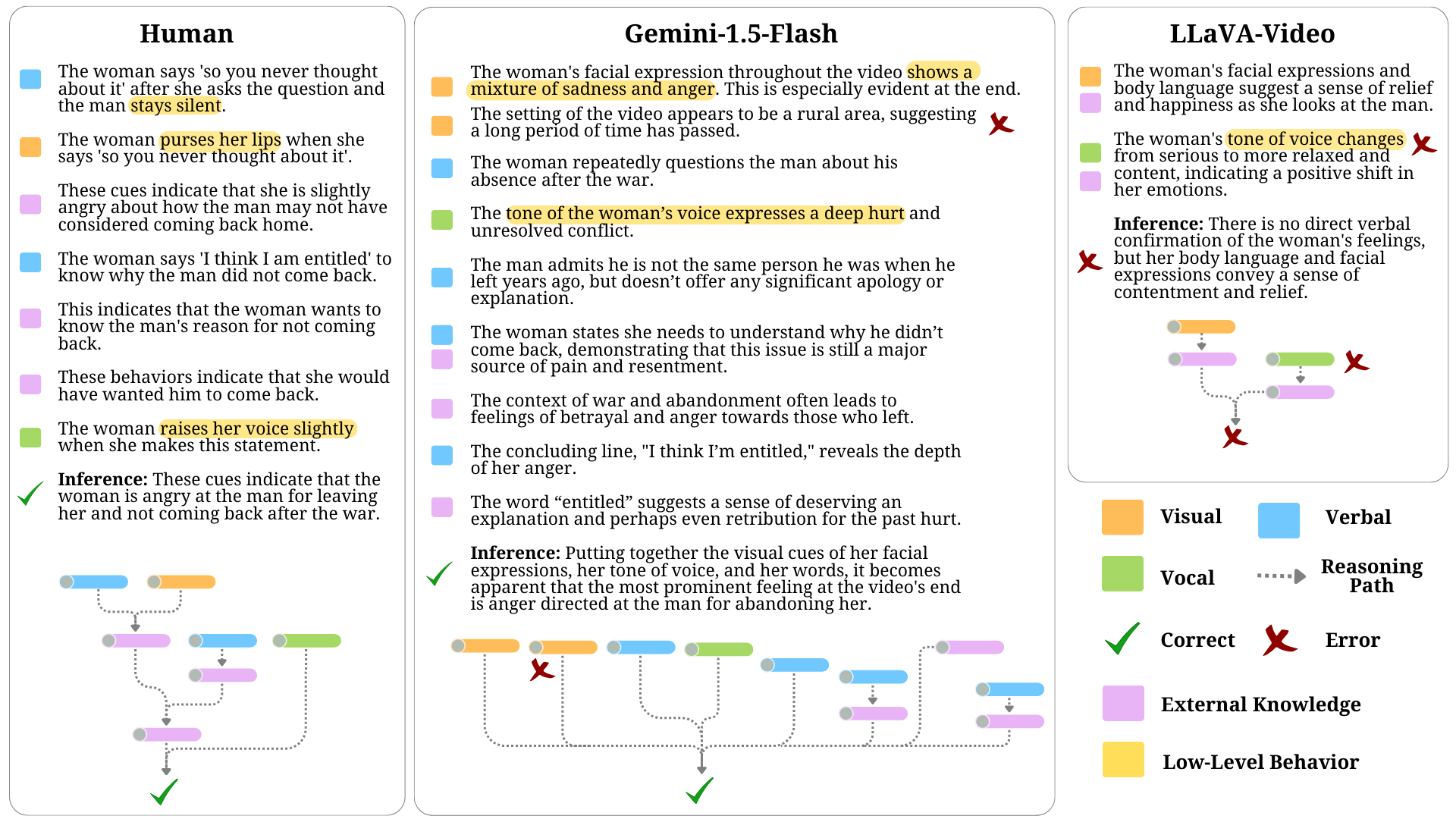}
    \caption{Representative social reasoning traces from a Human Annotator, Gemini-1.5-Flash, and LLaVA-Video. These examples illustrate the grounded social reasoning abilities and reasoning structures across humans and models.}
    \label{fig:qualitative}
\end{figure*}

\paragraph{Fine-Grained Multimodal Grounding} The final 6 metrics study evidence referenced by model reasoning traces. The \textbf{AllModality Steps} metric serves as a proxy for how well models refer to fine-grained, multimodal cues and external knowledge. As seen in AllModality Steps results, \textit{all models referenced fewer pieces of multimodal evidence and external knowledge than human reasoning.}

While we discuss modality-specific findings below, our use of \name\ to study SOTA models is currently limited by their varying abilities. The Gemini-1.5-Flash API reports that audio is processed in-parallel on the backend, but other models do not process audio. However, despite the absence of audio, models referenced verbal and vocal evidence \textit{inferred} from visual frames. For example, LongVA generated vocal evidence about a woman's "tone suggesting frustration," after generating visual evidence about the woman's face appearing dissatisfied. As new models are developed in the coming years with abilities to jointly process video and audio, \name\ will continue to be applicable to study model abilities in grounded multimodal social reasoning.

\textbf{Visual Steps:} \textit{Closed-source models exhibited a strong ability to reference visual evidence}, with GPT-4o and Gemini-1.5-Flash closer to ground truth than open-source models. Performance on this metric for all models was highest at $k$ = 0. 

\textbf{Verbal Steps:} \textit{The ability of models to reference verbal evidence showed substantial variation}. Gemini-1.5-Flash exhibited the strongest ability to reference verbal cues (0.90 at $k=0$).
Model performance on this metric did not improve as $k$ increased, and GPT-4o referenced substantially fewer verbal cues in comparison to other models. 

\textbf{Vocal Steps:} \textit{The ability of models to reference vocal evidence was substantially lower than human ability}. LongVA referenced more vocal evidence than other models at $k$ = 0. Model performance on this metric did not improve as $k$ increased. 

\textbf{External Knowledge Steps:} \textit{The ability of models to reference external knowledge in social reasoning traces was lower than humans}. In contrast to trends observed in other modality step metrics, we find that providing additional few-shot samples improved the ability of Gemini-1.5-Flash, GPT-4o, LongVA (up to $k$ = 4), and LLaVA-Video (up to $k$ = 4) to reference external knowledge. These findings suggest that human social reasoning traces can be used to condition models to ground social reasoning with external knowledge references.  

\textbf{Num Steps}: \textit{Model reasoning traces varied in length and contained more steps than human traces}. LLaVA-Video and LongVA generated model reasoning traces that were most aligned with the length of human traces. Providing additional few-shot samples improved the ability of models to align with human social reasoning trace length. 

\subsection{Human and Qualitative Evaluation}
\label{sec:human}
We conducted human evaluation of model reasoning traces. Trained annotators analyzed 48 samples from all models, with model names and $k$ values  anonymized. Traces were rated to assess references to low-level cues (\textit{fine-grained}), information cross-referenced across steps (\textit{compositional}), relevant evidence (\textit{comprehensive}), \textit{correctness} of modality tags, and \textit{validity} of reasoning (details in Appendix \ref{subsec:qual_results}). Annotator agreement was computed with Cohen's Kappa $\kappa$ \cite{cohen1960coefficient}.

Reasoning traces from Gemini-1.5-Flash and GPT-4o received the highest ratings for \textit{fine-grained} ($\kappa$ = 0.87), \textit{compositional} ($\kappa$ = 0.92), \textit{comprehensive} ($\kappa$ = 0.94), and \textit{valid} ($\kappa$ = 0.94) reasoning, followed by LLaVA-Video and LongVA, with VideoChatGPT and VideoChat2 rated lowest. These findings validate model performance trends in Section \ref{subsec:quant}. Modality tag correctness across samples was 98\%, and metrics correlated with human judgements ($R^2 >$ 0.75 for Gemini-1.5-Flash semantic metrics, details in Appendix \ref{subsec:qual_results}), supporting the validity of benchmark processing. 

\paragraph{Evidence and Error Propagation} Figure \ref{fig:qualitative} illustrates the strong ability of Gemini-1.5-Flash to reference and integrate multimodal cues and external knowledge. In contrast, LLaVA-Video did not reference low-level cues and based its reasoning upon an incorrect premise which led to an incorrect inference. The human trace referred to \textit{fine-grained behaviors} (e.g., lip movements) that are \textit{not} present in the model traces, yet do influence the scene interpretation. Metrics in Section \ref{subsec:metrics} serve as proxies to estimate these types of information gaps between human and model reasoning. Our findings motivate future work on model training and architectures that better capture fine-grained cues and handle error propagation in reasoning.

\paragraph{Hierarchical Social Reasoning} Figure \ref{fig:qualitative} shows the \textit{hierarchical structure} of human social reasoning traces, in which low-level cues (e.g., brief lip movement) are referenced, combined, and re-interpreted as \textit{intermediate evidence} for further reasoning.  ``Forking" reasoning structures are common for humans interpreting everyday situations, unlike the linear ``long chain" structures for formal reasoning in domains such as mathematics \cite{perkins1989reasoning, galotti1989approaches}. Compared to human traces, model traces show flatter structures which have the potential to overlook intermediate evidence. Our findings motivate future work to train models capable of more hierarchical social reasoning.

\section{Conclusion}
\label{sec:conclusion}

We introduce \name, the first benchmark for fine-grained, grounded social reasoning abilities of multimodal models. Reasoning traces contributed by \name\ include multimodal cues and external knowledge concepts that humans find useful when performing social inferences. We define metrics to assess semantic and structural aspects of reasoning traces and contribute novel insights regarding gaps and opportunities to improve the grounded social reasoning capabilities of multimodal models. Future AI systems reasoning about social interactions must be able to \textit{ground} reasoning in concrete multimodal evidence and external knowledge concepts. \name\ serves as a first step towards studying and advancing this form of reasoning in AI systems.

\section{Limitations}
\label{sec:limitations}

\paragraph{Social Reasoning in Natural Language} The current scope of \name\ focuses on studying model-generated social reasoning traces in \textit{natural language}. This scope is necessary and relevant to contexts that require AI systems to generate natural language explanations of social inferences -- for example, a healthcare agent or hospital robot reasoning about human nonverbal behaviors during a nurse-patient social interaction. However, an open question remains regarding the extent to which natural language can effectively represent the nuances of human social interactions and social reasoning \cite{mathur-etal-2024-advancing}. It is possible that both humans and models verbalizing social reasoning through natural language are not fully  capturing \textit{why} they came to certain inferences \cite{turpin2023language}. Several lines of work in reasoning have operated in the \textit{latent space} of models instead of natural language \cite{hao2024training, geiping2025scaling}, and we believe \name\ informs and motivates future work to develop techniques to study social reasoning in the latent space. 

\paragraph{Video Lengths} The videos in \name\ each have a length of $\sim$1 minute, consistent with the lengths of existing video understanding benchmarks (e.g., \textsc{Social-IQ 1.0} and \textsc{Social-IQ 2.0} have 1-minute samples \cite{wilf2023social, zadeh2019social}, TVQA has 1.3 minute samples \cite{lei-etal-2018-tvqa}, and MEmoR has 30 second samples \cite{shen2020memor}). \textit{Social reasoning regularly occurs in micro-social and shorter-term contexts}; humans make split-second inferences about emotions \cite{nook2015new}, social behaviors and gestures \cite{beattie1994gestures}, and personality \cite{lin2021four}, among other social phenomena. The interactions in \name\ videos contain rich, nuanced social signals and multimodal behavioral dynamics that require social reasoning to interpret. Our paper demonstrates that current state-of-the-art models struggle to interpret 1-minute social interactions.  The length of our videos is not, in itself, a technical limitation of our research; however, we would like to motivate the need for community-driven curation of longer-form social interaction datasets in future years. 

\paragraph{Scope of the Study} Videos in \name\ have interactions in English, and annotators were required to be proficient in English. The study was scoped within these constraints, consistent with prior multimodal video understanding tasks (Table \ref{tab:social_benchmarks}). A multilingual and multicultural data collection was not within the scope of this research. Our paper motivates future research in multimodal social reasoning that includes a community-driven curation of interaction data across sociocultural contexts. 







\section{Ethics}
\label{sec:ethics}

\paragraph{Ethical Annotation} We curated  annotations from videos in existing publicly available datasets. We hired workers from Prolific to annotate reasoning traces. All workers received fair compensation for their annotation (\$12 per hour, pro-rated). Worker privacy and confidentiality were respected, with no identifiable information stored. Further details on Prolific annotation are in Appendix \ref{subsec:annotation_appendix}.

\paragraph{Bias Considerations} Annotators in the original \textsc{Social-IQ 2.0} dataset, from which we sourced seed videos, used terms such as "man" in alignment with annotator perception of gender. In \name\ we did not frame judgements about gender identity of individuals based on these annotations.  In \name, 45\% of samples refer to women, and 17\% make no reference to gender. Samples involving women do not have reasoning traces that refer more frequently to emotion words ($r$ = 0.033). We find no substantial difference in model performance across gender; for example, Gemini-1.5-Flash social inference accuracy is 74.9\% for samples solely involving women, 73.4\% for sampling solely involving men, 73.6\% for samples referring to multiple genders, and 77\% for samples that do not specify gender.

\paragraph{Environmental Statement} Experiments used a single A100 GPU, a carbon footprint of 1.24 kgCO2e, and an energy consumption of 3.72 kWh\footnote{http://calculator.green-algorithms.org}. 

\paragraph{Risks for Social Reasoning in AI}

Social reasoning abilities are essential for future AI systems to effectively work \textit{with} and \textit{alongside} humans. \name\ has the potential to support the research community in studying and advancing these capabilities in AI systems. We envision AI systems using social reasoning to enhance human autonomy, health, and well-being. However, these technologies exist with potential risks in amplifying toxicity \cite{zhou-etal-2023-cobra}, surveillance, and manipulation. We support broader research and policy efforts to mitigate against misuse and potential harms of socially-intelligent AI.

\section*{Acknowledgments} 
Leena Mathur is supported by the NSF Graduate Research Fellowship Program under Grant No. DGE2140739. This material is based upon work partially supported by National Institutes of Health awards R01MH125740, R01MH132225, and R21MH130767.  Any opinions, findings, conclusions, or recommendations expressed in this material are those of the authors and do not necessarily reflect the views of the sponsors, and no official endorsement should be inferred. Figure \ref{fig:overview} includes icon material available from https://icons8.com. 

\bibliography{custom}

\newpage
\clearpage
\appendix

\section{\name\ Dataset Appendix}
\label{sec:data_appendix}

\subsection{\name\ Data Sourcing}
\label{subsec:sourcing}
All videos in \name\ were sourced from publicly-available \textsc{Social-IQ 2.0} modeling challenge \cite{wilf2023social}. We obtained permission from the authors to access the original \textit{test set} videos and question-answer tuples from this dataset (275 videos and 1514 QA Tuples). Our use of these videos aligns with Social-IQ 2.0 repository's MIT license and intended research purpose. We chose to use these test set videos as candidate seed videos to build \name\ because the answers to questions posed about these videos \textit{have not been released online}, reducing chances of benchmark contamination \cite{xu2024benchmark}. At the current stage, we plan to avoid benchmark contamination in \name\ by not uploading human social reasoning trace annotations online, by using them for research purposes, and by maintaining a leaderboard for the community.  

We note that prior papers that test models on \textsc{Social-IQ 2.0} have used the validation set, not the test set \cite{xie2023multi, pirhadi2023just, guo2023desiq, li2024llms, agrawal2024listen, chen2024through}.  Therefore, we do not directly compare prior works' validation set performance with our results on the test set. 

During manual inspection of each video and QA tuple, we filtered out QA tuples with answer options containing ambiguity -- if at least 2 annotators judged a question to have ambiguous answer options (at least two plausibly-correct answers), we discarded the question. The resulting \name\ set contained 272 videos and 1486 QA tuples. These samples contain face-to-face dyadic and multi-party social interactions and questions that probe understanding of affective states, causal social dynamics, and social events. 

\subsection{\name\  Entities}
\label{subsec:data_info}

\begin{figure}[h]
    \centering
\includegraphics[width=1\linewidth]{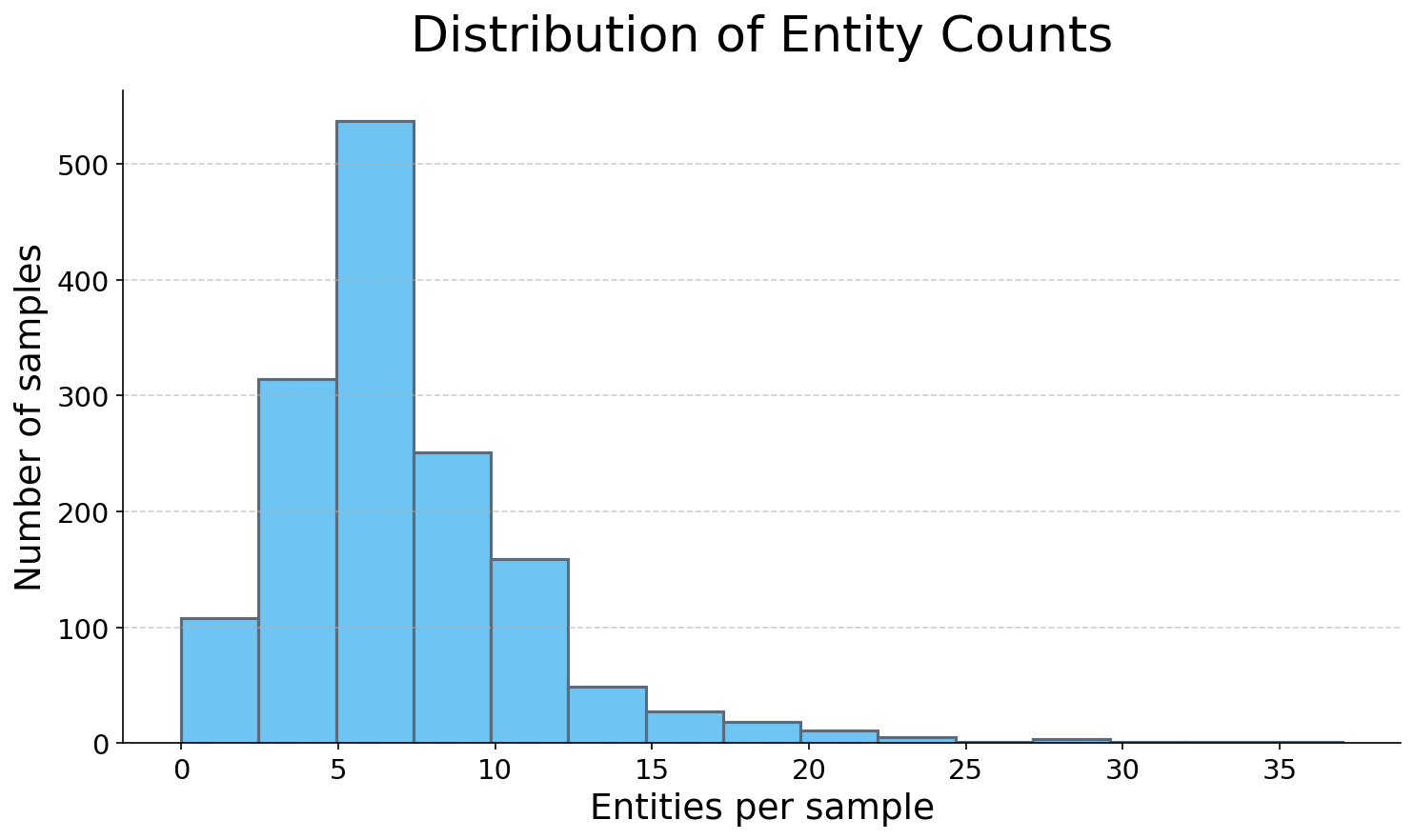}
    \caption{Distribution of entity counts in \name\ samples.}
    \label{fig:entity-distribution}
\end{figure}

\begin{figure}[h]
    \centering
\includegraphics[width=1\linewidth]{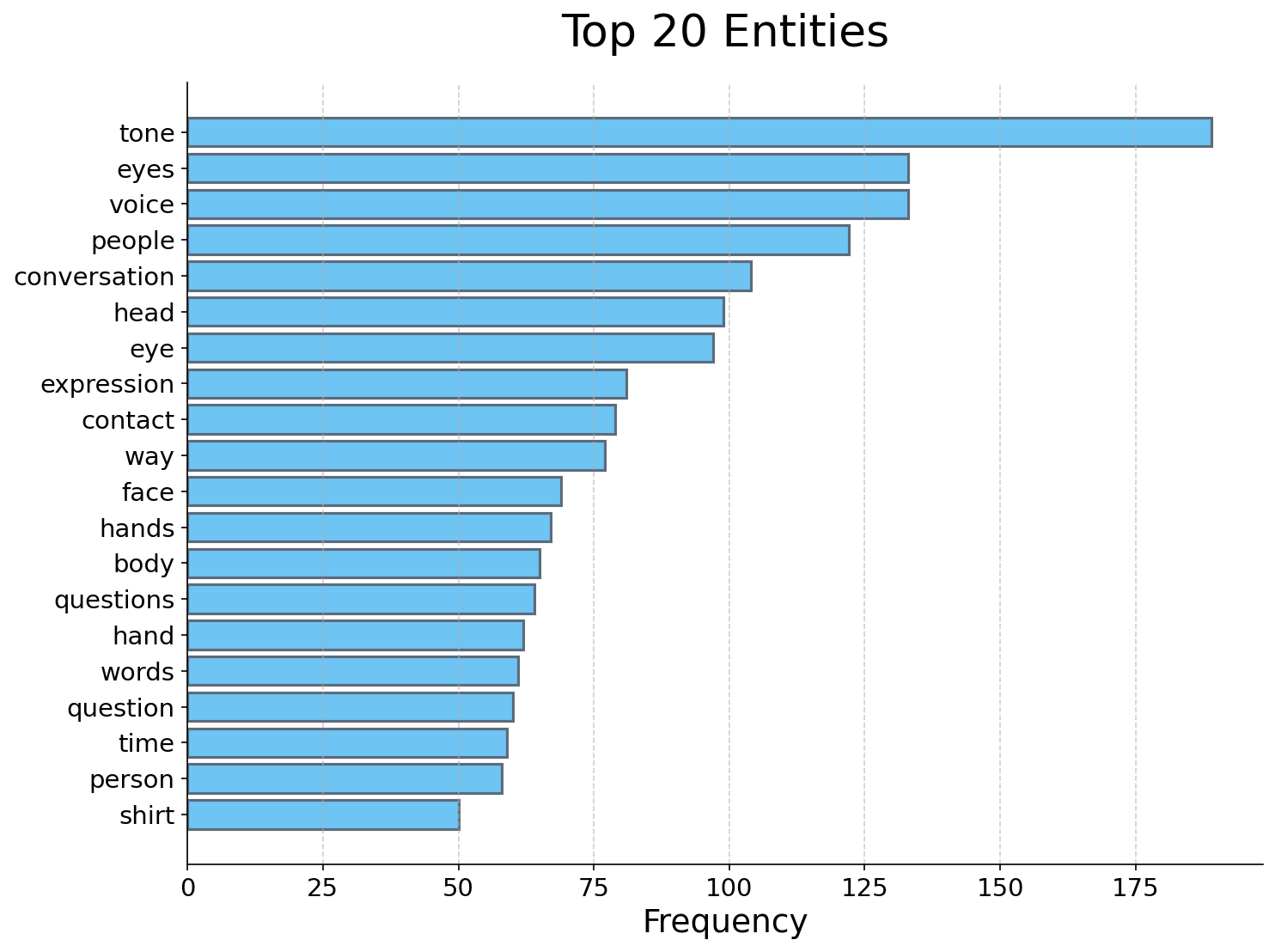}
\caption{Top entity counts mentioned by human annotators in \name\ reasoning traces.}
    \label{fig:entity-top}
\end{figure}

Across the 1486 samples and 11,253 entities (people, objects, concepts) mentioned in \name, we visualize the distribution of non-human entities (objects, concepts) in Figure \ref{fig:entity-distribution}, with an average of 7.6 unique entities referenced per human-annotated reasoning trace. Figure \ref{fig:entity-top} visualizes entities most frequently mentioned by human annotators (not including words such as "man" and "woman" repeated from question statements). As seen in Figure \ref{fig:entity-top}, human annotators constructing social reasoning traces focused on multimodal aspects of social interactions, with evidence spanning \textit{vocal} (e.g., "tone", "voice"), \textit{visual} (e.g., "eyes", "head", "body"), and \textit{verbal} (e.g, "words", "conversation") cues. These observations support the perspective that interpreting and reasoning about real-world social interactions requires an integration of multimodal information. 

Human annotators mentioned an average of 2.23 emotions per reasoning trace. Further analysis of the emotions referenced indicated that these emotions spanned a diverse range. Emotions frequently mentioned were the following: happy, serious, surprised, excited, nervous, calm, confused, angry, annoyed, comfortable, sarcastic, and positive. 

\subsection{\name\ Annotation }
\label{subsec:annotation_appendix}


Annotators were recruited on the Prolific\footnote{https://app.prolific.com} platform to perform an IRB-approved study with informed consent regarding our intended data use. Screens for annotators were employed to ensure that participants were adults, had access to a desktop to watch the videos and perform the annotation, were fluent in English, were based in the United States, and completed at least a high school education, and had high prior task approval rates on Prolific (97-100\% completion). Each annotator watched 2 videos and provided chains for all questions associated with each video (approximately 10 questions total per annotation). Instructions for annotators are listed in Figure \ref{fig:prolific}. Annotators were compensated at \$12 per hour pro-rated. This annotation process was time-intensive, with each human video annotation taking approximately $\sim$25 minutes. 

Before running the Prolific study, we ran 4 iterations of annotation instructions with pilot groups to refine instructions. We experimented with instructions that had annotators provide reasoning traces without knowing the correct answer and while knowing the correct answer. We found that providing annotators with the correct answer option did not change the detail, structure, or coherence of the reasoning traces generated. Therefore, we chose to provide annotators with the correct answer option (Figure \ref{fig:prolific}) and focus the large-scale Prolific data collection on obtaining detailed reasoning traces, instead of additional QA answer responses.  

\begin{figure*}[t]
\small
\centering
\begin{tcolorbox}[colframe=black!25, colback=white!95, coltitle=black, title=Annotation Instructions, width=0.95\textwidth, fonttitle=\bfseries]

In this study, you will be watching short videos that contain social interactions. You will be briefly explaining your thought process along several dimensions while answering questions about each video. Your answers will be entered through a Google spreadsheet.
Participants in this study must have the ability to watch videos and write logical text responses in English describing their step-by-step reasoning while interpreting videos. 

\vspace{0.5em}

If you give your consent to participate in this study, please select "I agree":  
\vspace{0.5em}
\texttt{<I agree>}

(If the annotator declines, they move to a page that states: ``As you do not wish to participate in this study, please return your submission on Prolific by selecting the 'Stop without completing' button.")

\vspace{0.5em}

In the provided spreadsheet (link below), you will see 2 videos linked in the "Video URL" column and a set of corresponding questions and answers for each video. The yellow-highlighted box indicates the correct answer for each question. Your goal is to do the following:

\begin{enumerate}
    \item Click the blue URL next to each set of questions in the "Video URL" column. Watch the video and pay attention to the behaviors of people interacting. 
    \item For each question associated with each video, provide the reasoning chain in order to answer the question:
    \begin{enumerate}
        \item Read the question, read the answer options, and note the correct answer (highlighted in yellow).
        \item In the "Evidence" column, describe the reasoning steps you took to reach the correct answer. Put each step in a separate row in the spreadsheet. (If applicable) steps should start with low-level observations (behavioral information from the video OR external knowledge) and build up to higher-level concepts.
    \end{enumerate}
    \item For each piece of evidence, in the same row, fill out the "Modality" column drop-down menu to indicate whether your evidence is from visual information (facial movements, gestures, anything vision-related, etc), vocal information (e.g., audio/speech), verbal information (e.g., the actual content of words being spoken), or external knowledge (e.g., cultural norms, your own understanding of what a behavior means).
    \item Please aim for 5-10 reasoning steps per question (if applicable).
\end{enumerate}

We will look at the spreadsheet during our manual review of submissions. All questions must have evidence presented with the answer selected: \texttt{<link to the spreadsheet>}

\end{tcolorbox}
\caption{Sample instructions provided to annotators on Prolific to provide \name\ reasoning traces.}
\label{fig:prolific}
\end{figure*}

\subsection{\name\ Validation}
\label{subsec:validation}
After obtaining human-annotated reasoning chains from Prolific, we conduct validation to ensure the validity and correctness of annotations. Two authors manually watched each video, read each question and corresponding set of answer options, and checked whether chains (1) represented valid reasoning paths, (2) had correct evidence tags for \textit{visual, verbal, vocal} and \textit{external knowledge}, and (3) comprehensively referred to relevant low-level social information in videos. If an annotator did not complete an annotation in a satisfactory manner (e.g., incomplete reasoning chain, minimal effort), the task was returned to them on Prolific. If annotations could be rapidly fixed (e.g., changing an incorrect modality tag from "visual" to "vocal"), the authors performed this fix themselves without sending the annotation task to the annotator. 

\subsection{\name\ Human Accuracy}
\label{subsec:human_acc}
For each video in \name, two human annotators on Prolific watched the video, read each question, and selected one of four provided answer options. Annotators were paid \$12 per hour (pro-rated) for completing each study. For each question, annotators also had the option to indicate whether or not they were ``uncertain" about the answer option they selected, and the authors examined these samples to ensure all QA tuples in \name\ had correct answers, to avoid the situation in which a sample could have more than one plausibly-correct answer option. Human annotators on Prolific achieved an accuracy of 85.3\% on the 1486 questions in \name. Inter-annotator agreement among Prolific annotators was positive (Cohen's $\kappa$ = 0.60) \cite{cohen1960coefficient}, and answer correctness was confirmed independently by authors, as described earlier. 

\section{\name\ Metrics Appendix}
\label{sec:metrics-appendix}

\subsection{Embeddings for Semantic Similarity} 
\label{subsec:semantic_embedding}

The metrics Similarity-Trace, Similarity-Step, and Similarity-Num Steps are computed with embeddings from the \texttt{all-MiniLM-L6-v2} from Sentence-BERT \cite{reimers-2019-sentence-bert}. We found this embedding strong, efficient, and useful for our task; as future embedding models are enhanced and released in the coming years, the \name\ framework allows for the embedding model called by semantic similarity metrics to be updated. 

\subsection{DifferenceSequence Metric}
\label{subsec:diffseq}
For model reasoning chain \(R_M\) with modality sequence \(S_M\) and human reasoning chain \(R_H\) with modality sequence \(S_H\), the DifferenceSequence ($DS$) metric is computed as a normalized similarity score by adapting the Levenshtein  distance \cite{levenshtein1966binary}  between \(S_M\) and \(S_H\):

\[
DS = 1 - \frac{\text{Levenshtein}(\mathit{S_M}, \mathit{S_H})}{|\mathit{S_M}| + |\mathit{S_H}|}
\]
\[
\text{Levenshtein}(S_M, S_H) \text{ is the following:}
\hspace*{5cm} 
\]
\[
= \begin{cases}
|S_M|, \small \text{if } |S_H|=0\\
|S_H|, \small \text{if } |S_M|=0\\ 
\min \Big[
\text{Levenshtein}(S_M[:-1], S_H[:-1]) + \delta, \\
\hspace{1em}\text{Levenshtein}(S_M[:-1], S_H) + 1, \\
\hspace{1em} \text{Levenshtein}(S_M, S_H[:-1]) + 1
\Big] \small\text{otherwise}
\end{cases}
\]
with \(\delta = \mathbbm{1}(S_M[-1] \neq S_H[-1])\), where \(\mathbbm{1}(\cdot)\) returns \(1\) if the final elements differ and \(0\) otherwise. 
We use an implementation\footnote{https://rapidfuzz.github.io/Levenshtein/index.html} that treats a substitution as equivalent to one insertion plus one deletion, making the distance effectively an InDel distance. We compute the minimum number of edits  that are needed to transform one sequence to another. Higher edit distances indicate that more edits are needed to align sequences (more dissimilarity).

Therefore, the overall \(DS\) similarity metric between \(S_M\) and \(S_H\) can range from 0 (maximum number of edits required) to 1 (minimum number of edits required). Higher \(DS\) values indicate greater structural similarity between the sequences \(\mathit{S_M}\) and \(\mathit{S_H}\), with respect to the type and order of modality evidence being referenced.

\begin{figure*}
\begin{tcolorbox}[colframe=black!25, colback=white!95, coltitle=black, title=\centering Model Prompt Template, width=\textwidth, fonttitle=\bfseries] 

Trace: Having a model provide a \textit{reasoning trace} in the response (remove the ``few-shot examples" for zero-shot experiments at $k$ = 0):

\begin{quote}
\texttt{Watch the video and answer questions about social information in the video, following the format of the examples below.}

\texttt{<FEW-SHOT EXAMPLE with Question and Answer Options>}

\texttt{Which of these is the correct answer? Output either A, B, C, or D.} \texttt{Provide the reasoning steps you took to get to this answer. Give 5 to 10 reasoning steps in bullet point form. Tag each bullet point step as visual, vocal, and verbal modality from the frames or external knowledge.  Use the format: The correct answer <insert answer>.The correct answer is <FEW-SHOT RESPONSE WITH ANSWER AND REASONING TRACE>.} 

\texttt{Which of these is the correct answer?  Output either A, B, C, or D.}  \texttt{<CURRENT QUESTION AND ANSWER OPTIONS> Provide the reasoning steps you took to get to this answer. Give 5 to 10 reasoning steps in bullet point form. Tag each bullet point step as visual, vocal, and verbal modality from the video or external knowledge.  Use the format: The correct answer is <insert answer>.} 
\end{quote}

\vspace{4mm}

No Trace: Having a model answer the question with \textit{no} reasoning trace (remove the ``few-shot examples" for zero-shot experiments at $k$ = 0): 

\begin{quote}
\texttt{Watch the video and answer questions about social information in the video, following the format of the examples below.}

\texttt{<FEW-SHOT EXAMPLE with Question and Answer Options>}

\texttt{Which of these is the correct answer? Output either A, B, C, or D. Use the format: The correct answer <insert answer>.The correct answer is <FEW-SHOT CORRECT ANSWER, FEW-SHOT REASONING CHAIN>.} 

\texttt{Which of these is the correct answer?  Output either A, B, C, or D.}  \texttt{<CURRENT QUESTION AND ANSWER OPTIONS> Use the format: The correct answer is <insert answer>.} 
\end{quote}

Note: For GPT-4o prompts, we found that we needed to replace the word \textit{video} with \textit{frames} to avoid  error messages associated with video processing (e.g., ``\textit{I am unable to view or analyze video frames directly. However, I can help answer questions based on descriptions or provide general information. Let me know how I can assist you!}"). Other models did not have this issue. 

\vspace{4mm}


\vspace{4mm}

\end{tcolorbox}
\caption{Information on model prompts to obtain reasoning traces and inferences from models.}
\label{fig:model_prompts}
\end{figure*}

\subsection{Emotion Named Entity Recognition}
\label{subsec:emotionmetric}
We perform NER with spaCy \cite{Honnibal_spaCy_Industrial-strength_Natural_2020}. In spaCy's NER v3 configuration, we broadly defined an \texttt{emotion} label as a \texttt{``description of how a person is feeling"}. To avoid identifying words like "feels" as entities, we also passed an example into the spaCy NER configuration that explicitly labeled "feels" as not an emotion entity (e.g., \texttt{``She feels sad because her friend didn't come with her"}). There are an average of 2.23\small$\pm$\normalsize1.63 emotion entities referenced per human reasoning chain.

\subsection{Chain Processing for Metrics}
\label{subsec:chain_processing}
Computing metrics requires a standardized format for model-generated reasoning chains. Several model generations (in particular, the  generations from open-source models) required processing before metrics could be computed.  

We first parsed generations from models by splitting each generation based on its structure, such as the presence of line breaks, numbering, or sentences. For example, if a model generated a multi-sentence response, but did not include  line breaks or numbering within the response, we would split this model output by individual sentences (e.g., splitting on the  "." character).

We, then, parsed through each step and remove any steps that simply repeated the question or answer choices. We also removed phrases such as "reasoning step", "reasoning", or "the correct answer", as those phrases were often in steps like "The correct answer is A." or "Below are the reasoning steps:". In addition, during in-context learning experiments, we found that some model generations (in particular, GPT) would contain repetitions of sample chains within the model's response, leading the response to contain 2-3 reasoning chains. We automatically processed these responses by only taking the final reasoning chain out of these multiple chains and checking that this final chain was answering the original question. 

Models were tasked with tagging modalities for each step of their generated reasoning chains. To validate this process, we automatically checked whether each step included \textit{visual, vocal, verbal} or \textit{external knowledge} tags. However, models sometimes failed to tag modalities for their chains. To automatically handle these cases, we employed \texttt{GPT-4o-mini} \cite{hurst2024gpt} to tag modalities. We note that the models that needed this additional validation step were VideoChat2, VideoChatGPT, and LLaVA-Video; generations from these models did not have any modality tagging for the majority of their chains. The authors manually inspected a subset of GPT-generated modality tags for model-generated reasoning traces to verify accuracy. 

\section{\name\ Model Appendix}
\label{sec:models-appendix}

\subsection{Model Information}
\label{subsec:model_info}

Experiments were conducted with multimodal models that 
were selected for their SOTA performance on various video understanding tasks and have the ability to take a full video as input (2 closed-source models and 5 open-source models): Gemini 1.5 Flash \cite{team2024gemini}, GPT-4o \cite{gpt4o},
VideoChat2 \cite{li2023videochat}, Video-ChatGPT \cite{maaz2023video}, LLaVA-Video \cite{zhang2024video}, LLaVA-Video-Only \cite{zhang2024video}, LongVA \cite{zhang2024longva}. We summarize the models below, and Appendix Table \ref{tab:model_comparison_transposed} lists characteristics of these models: context length, tokens per frame, training max frame, parameter count, and backbone. Figure \ref{fig:model_prompts} describes the prompts given to models. Our benchmark allows experiments with any model that processes multimodal language and video input and outputs text, allowing \name\ to be used as a benchmark over time to study social reasoning. Experiments were conducted with one A100 GPU. 

\begin{table*}[ht!]
\centering
\scriptsize 
\begin{tabular}{@{}lccccccc@{}}
\toprule
\textbf{Model Property} & \textbf{VideoChat2}  & \textbf{Video-ChatGPT} & \textbf{LLaVA-Video} & \textbf{LLaVA-Video-Only} & \textbf{LongVA} & \textbf{GPT-4o} & \textbf{Gemini 1.5} \\ \midrule
\textbf{Context} & 2K & 2K  & 32K & 32K & 224K & 128K & 1M \\
\textbf{Max Frames} & 16 & 100 & 110 & 110  & 2000 & - & - \\
\textbf{Parameters} & 7B & 7B & 7B & 7B & 7B & - & - \\
\textbf{Backbone} & Vicuna-v0 & Vicuna-1.1 & Qwen2 & Qwen2 & Qwen2-Extended & - & - \\
\bottomrule
\end{tabular}
\normalsize
\caption{Information about models tested on \name: VideoChat \cite{li2023videochat}, VideoChat-GPT \cite{maaz2023video},  LLaVA-Video \cite{zhang2024llavanext-video}, LLaVA-Video-Only \cite{zhang2024llavanext-video}, LongVA \cite{zhang2024longva}, GPT-4o \cite{gpt4o}, Gemini 1.5 Flash \cite{team2024gemini}. All information is completed based on current public reports and repositories.}
\label{tab:model_comparison_transposed}
\end{table*}

\paragraph{VideoChat2} The VideoChat2 model \cite{li2024videochat2} has an architecture with UMT-L vision encoder \cite{li2023unmasked}, QFormer and Vicuna-7B v0 language model, has 7B parameters, can process up to 16 frames, and was trained with instruction tuning on a collection of 34 tasks spanning conversations, captions, visual question-answering, reasoning, and classification.

\paragraph{Video-ChatGPT}
The VideoChat-GPT model \cite{maaz2023video} has an architecture built on top of LLaVA, with a CLIP vision encoder \cite{radford2021learning} and a Vicuna-7B v1.1 language model. The VideoChat-GPT model can process up to 100 frames, and was trained on video instruction pairs from the VideoInstruct100K dataset. 


\paragraph{LLaVA-Video}
The LLaVA-Video model \texttt{LLaVA-Video-7B-Qwen2} \cite{zhang2024video}
has an architecture with a SigLIP SO400M vision transformer and Qwen2 language model, has 7B parameters, can process up to 110 frames, and was trained on mixture of single image, multi-image, and video tasks from the LLaVA-Video-178K and LLaVA-OneVision datasets \cite{li2024llava}. 

\paragraph{LLaVA-Video-Only} The LLaVA-Video-Only model \texttt{LLaVA-Video-7B-Qwen2-Video-Only} is identical to the LLaVA-Video model, with the exception of the training data \cite{zhang2024video}. LLaVA-Video-Only was solely trained on the LLaVA-Video-178K dataset.  

\paragraph{LongVA} The LongVA model \texttt{LongVA-7B-DPO} \cite{zhang2024longva} aligns a unified multimodal transformer (UMT) with QFormer and aligns this visual encoder with a Qwen2 7B language model. LongVA was trained on visual instruction-following datasets and multimodal document data and has a context length of over 200,000 visual tokens; this longer context length was  achieved by extending the context length of the language backbone to train on long text samples, before performing multimodal alignment and additional training to transfer this  ability to the multimodal domain. 


\paragraph{GPT-4o} The GPT-4o model is a closed-source  model from OpenAI \cite{hurst2024gpt}. The technical report for GPT-4o refers to this model as ``omnimodal" with the ability to accept inputs from text, audio, image, and video and generate outputs with text, audio, and image. The API access to this model supports video frame inputs and text inputs. 

\paragraph{Gemini-1.5-Flash} The Gemini-1.5-Flash model is a closed-source model from Google. The API access to this model supports video inputs and text inputs, up to a context length of approximately 1 million tokens. Gemini-1.5-Flash was distilled from the larger Gemini-1.5-Pro sparse mixture-of-experts transformer  \cite{team2024gemini}.

\subsection{Model Generation Notes}
\label{subsec:modelgen}

We note observations here on model generations. VideoChat generations for $k\in{0, 1, 2, 4}$ produced full sentences explaining reasoning traces, however the generation quality eroded for $k \in {8, 16}$. Several samples from these settings of $k$ were repeated words and short phrases (e.g.,`` \texttt{the the the...and and and and}").

Similarly, VideoChat-GPT generations for $k\in{0, 1, 2, 4}$ produced full sentences explaining reasoning, however the generation quality eroded for $k \in {8, 16}$. Samples from these settings of $k$ were repeated short words and letters  (e.g.,``\texttt{or or or, or}" and ``\texttt{B ( ( ( ( ( B}").

LLaVA-Video generations for $k\in{0, 1, 2, 4, 8, 16}$ produced full sentences explaining reasoning, however the generations as $k$ increased in $k\in{4, 8, 16}$ began to answer fewer questions. LLaVA-Video-Only answered questions, but did not generate reasoning traces; this model was discussed in Section \ref{sec:experiments} solely for the social inference accuracy metric.

These model generation challenges were not observed for LongVA or Gemini-1.5. GPT-4o initially generated ``\texttt{I'm sorry, I can't assist with that.}" as one of the reasoning trace steps for several questions, before answering the question. 

\section{Auxiliary Experiments}
\label{sec:extra_experiments}

\subsection{Social Inference Accuracy and Reasoning Trace Lengths} We hypothesized that models may perform worse on inferences that primarily rely on \textit{implicit} cues and \textit{contextual} knowledge. One proxy for this reliance is the length of a human trace -- if humans perform an inference immediately and only need to verbalize one reasoning step, that step was more likely to involve  implicit cues with contextual nuances (e.g., rapidly interpreting body language based on external knowledge). For samples with 1 reasoning step, 53\% referenced external knowledge in this first piece of evidence, in contrast to 33\% of samples with 5 reasoning steps and 20\% of samples with 10 reasoning steps. 

\begin{figure}[t]
    \centering
    \includegraphics[width=1\linewidth]{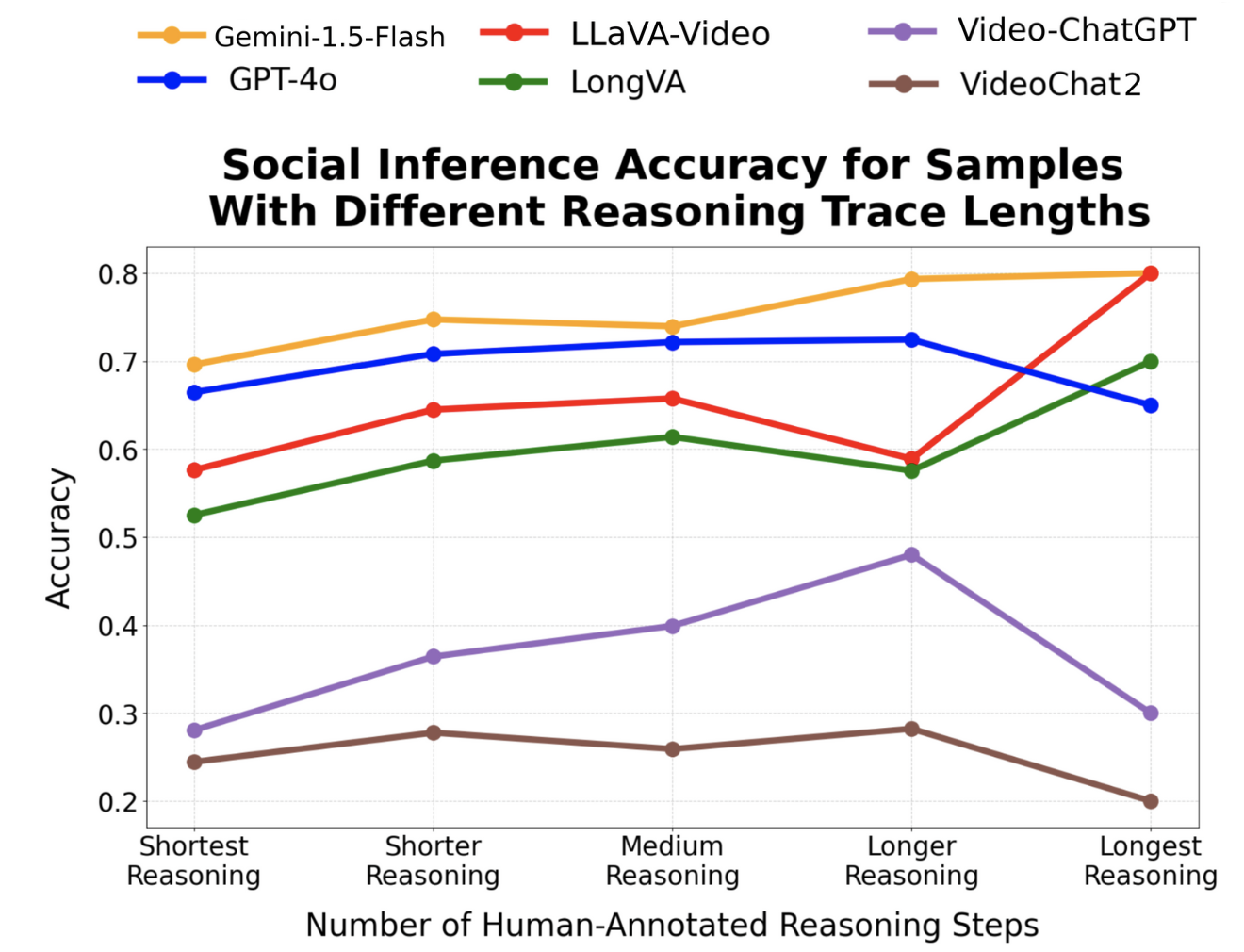}
    \caption{Social inference accuracy of models ($k$ = 0) across samples with different numbers of human-annotated reasoning steps (binned into quintile by reasoning trace length).}
    \label{fig:lengths}
\end{figure}

We examined model social inference performance  across samples with different lengths of human reasoning traces, visualized for $k$ = 0 in Figure \ref{fig:lengths}. 
\textit{Overall, multimodal models social inference performance was lower for samples with shorter human reasoning traces and higher for samples with longer human reasoning traces.} This trend was observed for both larger closed-source models and smaller open-source models that represent different training data distributions, architectures, and training techniques. For example, Gemini-1.5-Flash and LLaVA-Video achieved accuracies of 70\% and 58\%, respectively, for samples with the shortest reasoning traces and both achieved 80\% for samples with the longest reasoning traces. These results reinforce the perspective that current models (regardless of size) are not sufficient for strong multimodal social reasoning performance in domains requiring more contextual understanding.

\begin{figure}[t]
    \centering
    \includegraphics[width=1\linewidth]{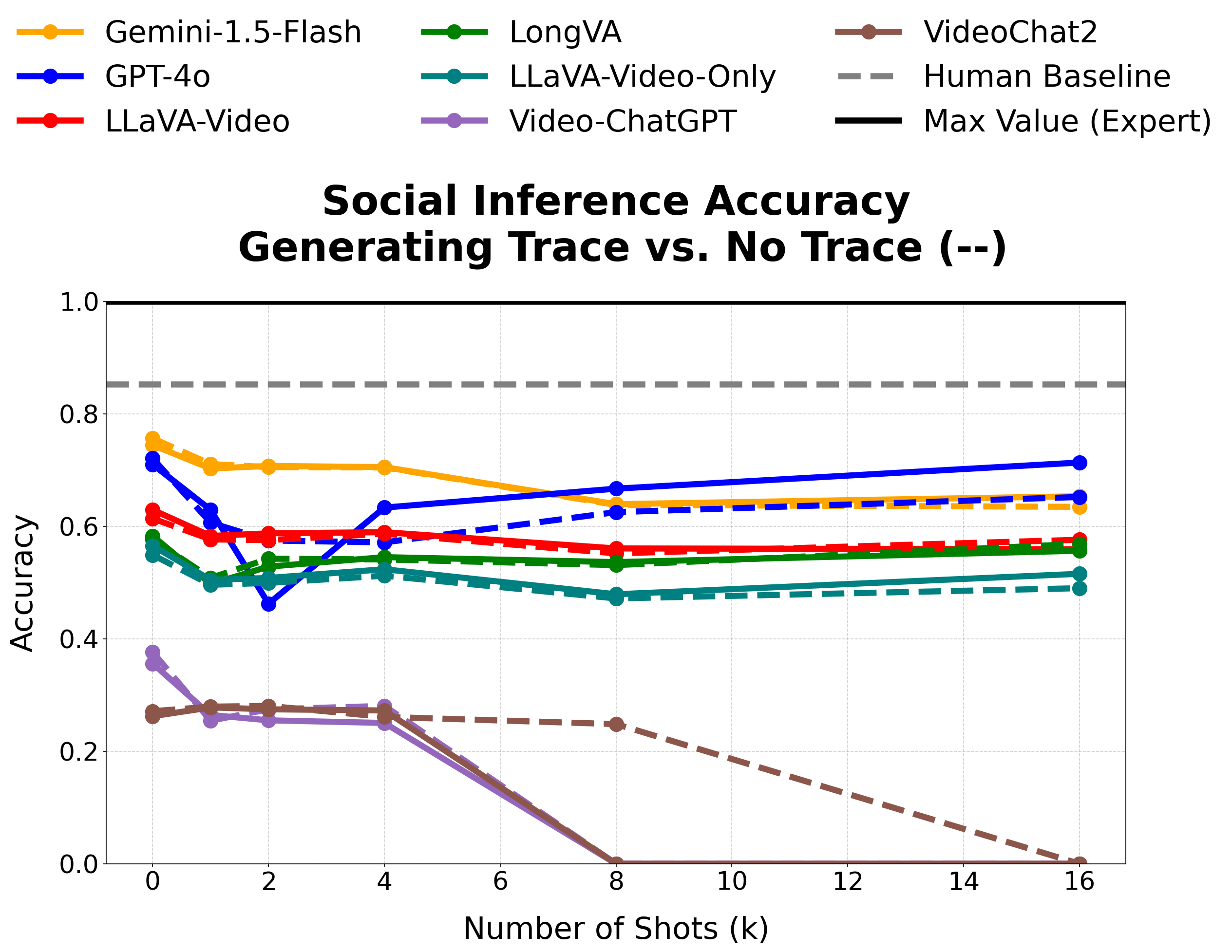}
    \caption{Social inference of models that generated reasoning traces with answers (Trace) compared to models that generated solely answers (No Trace), across different numbers of few-shot samples $k$.}
    \label{fig:cot}
\end{figure}
\raggedbottom


\begin{table*}[ht]
    \centering
    \small
    \caption{Human evaluation mean annotator scores for reasoning trace samples across models. }
    \label{tab:avg_scores}
    \resizebox{0.8\textwidth}{!}{%
    \begin{tabular}{lcccccc}
        \toprule
        \textbf{Model} & \textbf{Shot} & \textbf{Fine-Grained}& \textbf{Compositional} & \textbf{Modality Tags} & \textbf{Validity} & \textbf{Comprehensive} \\
        \midrule
        VideoChat2   & 0.0  & 1.50 & 1.00 & 0.5 & 0.0 & 1.25 \\
          & 1.0  & 1.00 & 1.00 & 1.0 & 0.0 & 1.00 \\
           & 4.0  & 1.00 & 1.00 & 1.0 & 0.0 & 1.00 \\
           & 16.0 & 1.00 & 1.00 & 1.0 & 0.0 & 1.00 \\
        \midrule
          Video-ChatGPT& 0.0  & 1.00 & 1.00 & 1.0 & 0.0 & 1.00 \\
        & 1.0  & 1.00 & 1.00 & 1.0 & 0.0 & 1.00 \\
        & 4.0  & 1.00 & 1.00 & 1.0 & 0.0 & 1.00 \\
        & 16.0 & 1.00 & 1.00 & 1.0 & 0.0 & 1.00 \\
        \midrule
        LongVA      & 0.0  & 2.75 & 2.75 & 1.0 & 1.0 & 4.50 \\
             & 1.0  & 3.00 & 3.25 & 1.0 & 1.0 & 3.75 \\
             & 4.0  & 2.25 & 3.50 & 1.0 & 1.0 & 3.50 \\
             & 16.0 & 2.25 & 2.75 & 1.0 & 1.0 & 3.25 \\
        \midrule  
          LLaVA-Video   & 0.0  & 3.00 & 2.50 & 1.0 & 1.0 & 4.00 \\
          & 1.0  & 2.00 & 1.75 & 1.0 & 0.5 & 3.00 \\
        & 4.0  & 1.00 & 1.00 & 1.0 & 0.0 & 1.00 \\
          & 16.0 & 1.00 & 1.00 & 1.0 & 0.0 & 1.00 \\
          \midrule
         GPT-4o     & 0.0  & 3.50 & 3.00 & 1.0 & 1.0 & 5.00 \\
               & 1.0  & 3.50 & 3.00 & 1.0 & 1.0 & 5.00 \\
                & 4.0  & 4.00 & 3.50 & 1.0 & 1.0 & 4.50 \\
                & 16.0 & 4.00 & 4.00 & 1.0 & 1.0 & 4.50 \\
        \midrule
        Gemini-1.5-Flash      & 0.0  & 4.25 & 4.25 & 1.0 & 1.0 & 4.75 \\
              & 1.0  & 4.00 & 2.50 & 1.0 & 1.0 & 4.75 \\
              & 4.0  & 3.75 & 3.75 & 1.0 & 1.0 & 3.00 \\
            & 16.0 & 4.25 & 4.25 & 1.0 & 1.0 & 4.50 \\
      
        \bottomrule
    \end{tabular}%
    }
\end{table*}

\subsection{Does Chain-of-Thought Prompting Elicit Multimodal Social Reasoning?}

Chain-of-thought (CoT) prompting \cite{wei2022chain} has been a prevalent approach to elicit model reasoning abilities in domains such as mathematics and code generation \cite{li2025structured}. We explore the effectiveness of this technique for multimodal social reasoning with \name. Figure \ref{fig:cot} visualizes social inference results for models under two settings: \textit{Trace} (models generate step-by-step CoT traces while answering the question) and \textit{No Trace} (models generate only answers and do not generate "step by step" traces). We test models  with zero-shot and few-shot settings, with $k \in$ \{0, 1, 2, 4, 8, 16\}. Prompts are described in Figure \ref{fig:model_prompts}. 

\textit{We find that CoT prompting does not substantially improve the social reasoning performance of models}, with the exception of GPT-4o (using CoT performs 6-8\% higher than without CoT after $k$=4). Unlike domains such as mathematics with more formal step-by-step reasoning paths, social reasoning often involves interpreting and integrating ambiguous, context-dependent cues across actors, modalities, and time \cite{mathur-etal-2024-advancing}. Social inference is a form of \textit{informal reasoning} that does not often verbalize as a ``chain-like" process \cite{galotti1989approaches}. Models with CoT prompting have been found to rely on task priors from pretraining distributions \cite{chochlakis2024larger}; it is possible these priors are less effective to elicit social reasoning, compared to other forms of reasoning.  





\section{Human Evaluation Details}
\label{subsec:qual_results}


Trained annotators analyzed 48 samples from all models at $k \in$ {0, 1, 4, 16}. The $k$ values and model identities associated with traces were anonymized before presenting samples to annotators in a spreadsheet to rate. For each reasoning trace, annotators watched the corresponding video, read the question, read the answer options, and read the correct answer, in order to rate the quality of the reasoning trace. Human evaluation was conducted to assess reasoning traces along the following dimensions:

\textbf{Fine-grained}: The extent to which reasoning traces were fine-grained was assessed on a scale of 1 to 5, with 1 for no references to low-level behavior and 5 for dense references (e.g., references to low-level behaviors in a majority of steps). 

\textbf{Compositional}: Compositionality in reasoning traces was assessed on a scale of 1 to 5, with 1 for minimal compositionality (no cross-references of information across reasoning steps) and 5 for high compositionality across steps. 

\textbf{Comprehensive}: The extent to which reasoning traces were comprehensive was assessed on a scale of 1 to 5, with 1 referring to  traces missing critical  information (not referencing relevant video content) and 5 being fully-comprehensive (capturing all relevant content towards an inference).

\textbf{Modality Tag Correctness}: The accuracy of modality and external knowledge tags for each of the reasoning steps within a given trace was assessed with a binary score (1 for correct, 0 for the presence of any error in the trace). 

\textbf{Validity of Reasoning}: The validity of the reasoning in a trace, referring to whether or not the trace represented logical combinations of information. It is possible for a reasoning trace to reference minimal low-level information, yet still represent valid reasoning (motivating the inclusion of this dimension). This dimension was given a binary rating (1 for valid, 0 for invalid). 

Raw averages from this human evaluation process are presented in Table \ref{tab:avg_scores}. We discuss findings from human evaluation in Section \ref{sec:human}. Annotator agreement was computed with Cohen's Kappa $\kappa$ \cite{cohen1960coefficient}. We found strong inter-annotator  agreement in ratings across dimensions: $\kappa$ = 0.87 for "fine-grained" ratings, $\kappa$ = 0.92 for "compositional" ratings, $\kappa$ = 0.94 for "comprehensive" ratings, and $\kappa$ = 0.94 for "validity of reasoning" ratings. The "modality tag correctness" ratings across samples was 98\% with errors specifically occurring in modality tags for VideoChat2 traces. 

We note that reasoning trace quality for LLaVA-Video, VideoChat2, and Video-ChatGPT, in particular, decreased at $k$ = 4. These findings from human evaluation are aligned with quantitative findings in Section \ref{sec:experiments} and subjective model generation observations in Appendix \ref{subsec:modelgen}. 

Our automated metrics yield similar model rankings to human evaluation and strong correlation to human judgements, supporting the reliability of these metrics as proxies for reasoning trace quality.  Gemini-1.5-Flash achieves the highest score in human judgements for referencing “Low-Level” behavioral cues and achieves the highest score in automated metrics such as Accuracy, overall semantic similarity (Similarity-Trace), and step-level semantic similarity (Similarity-Num Steps). Similarly, GPT-4o and LLaVA-Video, followed closely by LongVA, score higher than Video-ChatGPT and VideoChat2 on both automated metrics and human judgments. This rank-based alignment indicates that automated proxy metrics can capture reasoning quality signals that human evaluators identified.

These trends observed from rank-based alignment are supported by correlation analyses between model metrics and human judgments. For Gemini-1.5-Flash and LLaVA-Video (highest-performing closed-source model and open-source model), the Similarity-Trace and Similarity-Step metrics exhibit $R^2$ values of 0.75 and 0.61, respectively, demonstrating strong alignment with human judgments of how effectively models ground reasoning in low-level behavioral cues.  Strong correlations indicate that (1) these automated metrics can be viewed as effective predictors of low-level social reasoning quality and (2) there is room for future community research into automated evaluation of social reasoning quality. \name\ contributes a new benchmark and findings for the community to study this direction. 


\begin{table*}[h]
\centering
\scriptsize
\caption{Social Inference Accuracy on \textsc{Social-Genome} for Models Across Different Numbers of Shots ($k$)}
\label{tab:accuracy-metrics-comparison}
\resizebox{\textwidth}{!}{
\begin{tabular}{@{}llcccccc@{}}
\toprule
\textbf{Model} & \textbf{Chain Type} & \textbf{\textit{k} = 0} & \textbf{\textit{k} = 1} & \textbf{\textit{k} = 2} & \textbf{\textit{k} = 4} & \textbf{\textit{k} = 8} & \textbf{\textit{k} = 16} \\
\midrule

VideoChat2 & Chain & 0.2624 & 0.2779 & 0.2746 & 0.2725 & 0.0000 & 0.0000 \\
 & No Chain & 0.2712 & 0.2793 & 0.2806 & 0.2611 & 0.2483 & 0.0000 \\
\midrule

Video-ChatGPT & Chain & 0.3560 & 0.2645 & 0.2550 & 0.2503 & 0.0000 & 0.0000 \\
 & No Chain & 0.3762 & 0.2544 & 0.2746 & 0.2806 & 0.0000 & 0.0000 \\
\midrule

LongVA & Chain & 0.5828 & 0.4973 & 0.5283 & 0.5451 & 0.5363 & 0.5565 \\
 & No Chain & 0.5767 & 0.5081 & 0.5424 & 0.5410 & 0.5310 & 0.5686 \\
\midrule

LLaVA-Video & Chain & 0.6292 & 0.5828 & 0.5875 & 0.5895 & 0.5606 & 0.5592 \\
 & No Chain & 0.6137 & 0.5767 & 0.5754 & 0.5868 & 0.5518 & 0.5760 \\
\midrule

LLaVA-Video-Only & Chain & 0.5653 & 0.5047 & 0.5081 & 0.5236 & 0.4791 & 0.5155 \\
 & No Chain & 0.5491 & 0.4960 & 0.4993 & 0.5121 & 0.4717 & 0.4899 \\
\midrule

GPT-4o & Chain & 0.7100 & 0.6292 & 0.4623 & 0.6332 & 0.6669 & 0.7133 \\
 & No Chain & 0.7207 & 0.6063 & 0.5747 & 0.5713 & 0.6252 & 0.6521 \\
\midrule

Gemini-1.5-Flash & Chain & 0.7443 & 0.7026 & 0.7073 & 0.7052 & 0.6393 & 0.6534 \\
 & No Chain & 0.7564 & 0.7106 & 0.7052 & 0.7046 & 0.6380 & 0.6346 \\
\bottomrule
\end{tabular}
}
\end{table*}

\begin{table*}[ht]
\centering
\small
\caption{Model performance for semantic and structural similarity metrics across different numbers of few-shot samples $k$. These are the raw results (metrics visualized in Figure \ref{fig:overall} were normalized, as discussed in the metrics section). * refers to model reasoning traces that were less coherent (Appendix \ref{subsec:modelgen}). The highest-performance at each value  of $k$ is bolded.}
\label{tab:core-metrics}
\resizebox{\textwidth}{!}{
\begin{tabular}{@{}llcccccc@{}}
\toprule
\textbf{Metric} & \textbf{Model} & \textbf{\textit{k} = 0} & \textbf{\textit{k} = 1} & \textbf{\textit{k} = 2} & \textbf{\textit{k} = 4} & \textbf{\textit{k} = 8} & \textbf{\textit{k} = 16} \\
\midrule

\textbf{Similariy-Trace} & VideoChat2 & 0.4138 & 0.3954 & 0.4084 & 0.4079 & 0.0734 & 0.0597 \\
 & Video-ChatGPT & 0.4484 & 0.4484 & 0.4533 & 0.4354 & 0.0266 & 0.0760 \\
 & LongVA & 0.4533 & 0.4865 & 0.4929 & \textbf{0.4994} & \textbf{0.4994} & \textbf{0.4998} \\
 & LLaVA-Video & 0.4915 & 0.4193 & 0.4552 & 0.4178* & 0.4629* & 0.4731* \\
 & GPT-4o & 0.4631 & 0.4332 & 0.3437 & 0.4363 & 0.4479 & 0.4648 \\
 & Gemini-1.5-Flash & \textbf{0.5157} & \textbf{0.5060} & \textbf{0.5021} & 0.4891 & 0.4887 & 0.4850 \\
\midrule

\textbf{Similarity-Step} & VideoChat & 0.3864 & 0.3804 & 0.3856 & 0.3749 & 0.0909 & 0.0760 \\
 & Video-ChatGPT & \textbf{0.4524} & \textbf{0.4822} & \textbf{0.4836} & \textbf{0.4700} & 0.0578 & 0.1059 \\
 & LongVA & 0.3898 & 0.4148 & 0.4185 & 0.4231 & 0.4297 & 0.4310 \\
 & LLaVA-Video & 0.4462 & 0.4090 & 0.4330 & 0.4109 & \textbf{0.4641} & \textbf{0.4856} \\
 & GPT-4o & 0.4016 & 0.3540 & 0.2943 & 0.3574 & 0.3682 & 0.3837 \\
 & Gemini-1.5-Flash & 0.4340 & 0.4378 & 0.4311 & 0.4234 & 0.4231 & 0.4210 \\
\midrule

\textbf{Similarity-Num Steps} & VideoChat2 & 0.5168 & 0.4623 & 0.5983 & 0.4764 & 0.0000 & 0.0000 \\
 & Video-ChatGPT & 0.3267 & 0.2053 & 0.2144 & 0.1989 & 0.0000 & 0.0000 \\
 & LongVA & 0.4186 & 0.4529 & 0.4246 & 0.4179 & 0.4164 & 0.4181 \\
 & LLaVA-Video & 0.7719 & 0.6970 & \textbf{0.9205} & \textbf{0.7762}* & 0.4167* & 0.3750* \\
 & GPT-4o & 0.6226 & 0.4480 & 0.3496 & 0.4575 & 0.4440 & 0.4584 \\
 & Gemini-1.5-Flash & \textbf{1.0376} & \textbf{0.9144} & 0.7312 & 0.6984 & \textbf{0.6436} & \textbf{0.6525} \\
\midrule

\textbf{EmotionMetric} & VideoChat2 & 0.0754 & \textbf{0.2328} & 0.2530 & 0.2322 & 0.0034 & 0.0007 \\
 & Video-ChatGPT & 0.2358 & 0.1713 & 0.1904 & 0.0702 & 0.0000 & 0.0000 \\
 & LongVA & 0.3345 & 0.1306 & 0.3244 & 0.3075 & 0.3106 & 0.3121 \\
 & LLaVA-Video & 0.3579 & 0.1077 & 0.1038 & 0.2168* & 0.1667* & 0.2500* \\
 & GPT-4o & \textbf{0.6502} & 0.2005 & \textbf{0.6631} & \textbf{0.6725} & \textbf{0.6577} & \textbf{0.6286} \\
 & Gemini-1.5-Flash & 0.4959 & 0.1218 & 0.4152 & 0.3782 & 0.3557 & 0.3755 \\
\midrule

\textbf{DifferenceSequenceMetric} & VideoChat & 0.2673 & 0.2690 & 0.2838 &	0.2659 & 0.2141 &	0.2144 
\\
 & Video-ChatGPT & 0.2508 & 0.2352 &	0.2325 & 0.2335 &	0.2181 & 0.2111 \\
 & LongVA & 0.4266 &	0.4165 & 0.4102 &	0.4202 & 0.4045 &	0.3750 \\
 & LLaVA-Video & 0.3783 & 0.4819 & 0.4845 & 	0.4907* & 0.2589* &	0.1754* \\
 & GPT-4o & 0.4582 &	0.3956 & 0.3671 &	0.4167 & 0.4458 &	0.4638 \\
 & Gemini-1.5-Flash & \textbf{0.4591 }& \textbf{0.5301} & \textbf{0.5243} & \textbf{0.5312} & \textbf{0.5240} & \textbf{0.5295} \\
\bottomrule
\end{tabular}
}
\end{table*}

\begin{table*}[ht]
\centering
\small
\caption{Model Performance for metrics related to step-level evidence from modalities and external knowledge across different numbers of few-shot samples $k$. These are the raw results (metrics visualized in Figure \ref{fig:overall} were normalized, as discussed in the metrics section). * refers to reasoning traces that were less coherent (Appendix \ref{subsec:modelgen}). The highest-performance at each value  of $k$ is bolded.}
\label{tab:modality-metrics}
\resizebox{\textwidth}{!}{
\begin{tabular}{@{}llcccccc@{}}
\toprule
\textbf{Metric} & \textbf{Model} & \textbf{\textit{k} = 0} & \textbf{\textit{k} = 1} & \textbf{\textit{k} = 2} & \textbf{\textit{k} = 4} & \textbf{\textit{k} = 8} & \textbf{\textit{k} = 16} \\
\midrule

\textbf{All Modality Steps} & VideoChat2 & 1.2328 & 1.2436 & 1.3573 & 1.1817 & 0.6357* & 0.6184* \\
 & Video-ChatGPT & 0.8526 & 0.6801 & 0.6684 & 0.6638 & 0.6289* & 0.6519* \\
 & LongVA & 2.1406 & 1.7335 & 1.6669 & 1.6440 & 1.5659 & 1.4525 \\
 & LLaVA-Video & 1.6148 & 2.2242 & 2.0526 & 1.9231* & 0.7500* & 0.5625* \\
 & GPT-4o & 2.3245 & 1.7155 & 1.4566 & 1.7536 & 1.8676 & 1.9750 \\
 & Gemini-1.5-Flash & \textbf{2.4056} & \textbf{2.3806} & \textbf{2.3529} & \textbf{2.3146} & \textbf{2.2811} & \textbf{2.2798} \\
\midrule

\textbf{Visual Steps} & VideoChat2 & 0.6097 & 0.4105 & 0.5821 & 0.6326 & 0.0168* & 0.0000* \\
 & Video-ChatGPT & 0.2677 & 0.0919 & 0.0774 & 0.0755 & 0.0027* & 0.0000* \\
 & LongVA & 1.1299 & 0.9246 & 0.8863 & 0.8883 & 0.7782 & 0.7442 \\
 & LLaVA-Video & 1.0301 & 0.9367 & 0.9269 & 0.9580* & 0.0000* & 0.0000* \\
 & GPT-4o & \textbf{1.5894} & \textbf{1.4196} & 1.0725 & \textbf{1.4180} & \textbf{1.4922} & \textbf{1.5598} \\
 & Gemini-1.5-Flash & 1.3276 & 1.2806 & \textbf{1.1860} & 1.1395 & 1.0431 & 1.0513 \\
\midrule

\textbf{Verbal Steps} & VideoChat2 & 0.8338 & 0.8836 & 0.8358 & 0.8351 & 0.6316* & 0.6184* \\
 & Video-ChatGPT & 0.6533 & 0.5831 & 0.5774 & 0.5777 & 0.6030* & 0.6519* \\
 & LongVA & 0.8096 & 0.5942 & 0.6528 & 0.5989 & 0.6464 & 0.6988 \\
 & LLaVA-Video & 0.7643 & \textbf{0.7880} & 0.7487 & \textbf{0.8042}* & \textbf{0.9167}* & 0.6875* \\
 & GPT-4o & 0.7249 & 0.2811 & 0.3984 & 0.3057 & 0.3150 & 0.3320 \\
 & Gemini-1.5-Flash & \textbf{0.9316} & 0.7728 & \textbf{0.7709} & 0.7695 & 0.7360 & \textbf{0.7196} \\
\midrule

\textbf{Vocal Steps} & VideoChat2 & 0.1810 & 0.1635 & 0.2066 & 0.0740 & 0.0000* & 0.0000* \\
 & Video-ChatGPT & 0.0578 & 0.0202 & 0.0188 & 0.0117 & 0.0000* & 0.0000* \\
 & LongVA & \textbf{0.5162} & 0.2288 & 0.1844 & 0.1359 & 0.1365 & 0.1243 \\
 & LLaVA-Video & 0.2486 & 0.2375 & 0.1269 & 0.0490* & 0.0000* & 0.0000* \\
 & GPT-4o & 0.4302 & 0.1270 & 0.0589 & 0.1184 & 0.1270 & 0.1413 \\
 & Gemini-1.5-Flash & 0.4357 & \textbf{0.2984} & \textbf{0.2551} & \textbf{0.2291} & \textbf{0.2497} & \textbf{0.2579} \\
\midrule

\textbf{External Knowledge Steps} & VideoChat2 & 0.0229 & 0.3546 & 0.3230 & 0.2402 & 0.0054* & 0.0000* \\
 & Video-ChatGPT & 0.0130 & 0.0000 & 0.0000 & 0.0000 & 0.0232* & 0.0000* \\
 & LongVA & 0.3156 & 0.5565 & 0.5175 & 0.7174 & 0.6259 & 0.5044 \\
 & LLaVA-Video & 0.0792 & 1.4661 & 1.6128 & 1.5734* & 0.0000* & 0.0000* \\
 & GPT-4o & 0.6199 & 0.9108 & 0.7554 & 1.0402 & 1.1568 & 1.2366 \\
 & Gemini-1.5-Flash & \textbf{0.9856} & \textbf{1.4901} & \textbf{1.6190} & \textbf{1.6881} & \textbf{1.6854} & \textbf{1.6847} \\

\midrule
\textbf{Num Steps} & VideoChat2 & 4.9455 & 4.7840 & 5.8526 & 4.7638 & 3.1158 & 2.8876 \\
 & Video-ChatGPT & 3.4316 & 2.8955 & 2.8985 & 3.4564 & 2.8888 & 3.0443 \\
 & LongVA & 3.4233 & \textbf{2.3573 }& \textbf{2.1245} & \textbf{1.9455} & \textbf{1.8689} &\textbf{ 2.0475} \\
 & LLaVA-Video & \textbf{2.5574} & 4.0755 & 4.7333 & 4.4336* & 2.2500* & 2.5625* \\
 & GPT-4o & 4.0921 & 4.0757 & 3.5935 & 3.7699 & 3.7583 & 3.8039 \\
 & Gemini-1.5-Flash & 5.1060 & 4.2197 & 4.4624 & 3.9747 & 4.0445 & 3.6854 \\

\bottomrule
\end{tabular}
}
\end{table*}

\end{document}